\documentstyle[11pt,times,chicagor,smallsec,url]{article}

\newtheorem{THEOREM}{Theorem}[section]
\newenvironment{theorem}{\begin{THEOREM} \hspace{-.85em} {\bf :} }%
                        {\end{THEOREM}}
\newtheorem{LEMMA}[THEOREM]{Lemma}
\newenvironment{lemma}{\begin{LEMMA} \hspace{-.85em} {\bf :} }%
                      {\end{LEMMA}}
\newtheorem{COROLLARY}[THEOREM]{Corollary}
\newenvironment{corollary}{\begin{COROLLARY} \hspace{-.85em} {\bf :} }%
                          {\end{COROLLARY}}
\newtheorem{PROPOSITION}[THEOREM]{Proposition}
\newenvironment{proposition}{\begin{PROPOSITION} \hspace{-.85em} {\bf :} }%
                            {\end{PROPOSITION}}
\newtheorem{DEFINITION}[THEOREM]{Definition}
\newenvironment{definition}{\begin{DEFINITION} \hspace{-.85em} {\bf :} \rm}%
                            {\end{DEFINITION}}
\newtheorem{CLAIM}[THEOREM]{Claim}
\newenvironment{claim}{\begin{CLAIM} \hspace{-.85em} {\bf :} \rm}%
                            {\end{CLAIM}}
\newtheorem{EXAMPLE}[THEOREM]{Example}
\newenvironment{example}{\begin{EXAMPLE} \hspace{-.85em} {\bf :} \rm}%
                            {\end{EXAMPLE}}
\newtheorem{REMARK}[THEOREM]{Remark}
\newenvironment{remark}{\begin{REMARK} \hspace{-.85em} {\bf :} \rm}%
                            {\end{REMARK}}

\newcommand{\thm}{\begin{theorem}}
\newcommand{\lem}{\begin{lemma}}
\newcommand{\pro}{\begin{proposition}}
\newcommand{\dfn}{\begin{definition}}
\newcommand{\rem}{\begin{remark}}
\newcommand{\xam}{\begin{example}}
\newcommand{\cor}{\begin{corollary}}
\newcommand{\prf}{\noindent{\bf Proof:} }
\newcommand{\ethm}{\end{theorem}}
\newcommand{\elem}{\end{lemma}}
\newcommand{\epro}{\end{proposition}}
\newcommand{\edfn}{\bbox\end{definition}}
\newcommand{\erem}{\bbox\end{remark}}
\newcommand{\exam}{\bbox\end{example}}
\newcommand{\ecor}{\end{corollary}}
\newcommand{\eprf}{\bbox\vspace{0.1in}}
\newcommand{\beqn}{\begin{equation}}
\newcommand{\eeqn}{\end{equation}}

\newcommand{\bbox}{\vrule height7pt width4pt depth1pt}

\newcommand{\clm}{\begin{claim}}
\newcommand{\eclm}{\end{claim}}

\newcommand{\sat}{\models}

\newcommand{\stur}{\vdash}
\newcommand{\rimp}{\Rightarrow}

\newcommand{\dimp}{\Leftrightarrow}
\newcommand{\bor}{\bigvee}
\newcommand{\band}{\bigwedge}
\newcommand{\union}{\cup}
\newcommand{\inter}{\cap}

\renewcommand{\phi}{\varphi}

\newcommand{\A}{{\cal A}}

\newcommand{\C}{{\cal C}}

\newcommand{\F}{{\cal F}}

\newcommand{\K}{{\cal K}}
\newcommand{\M}{{\cal M}}
\newcommand{\N}{{\cal N}}

\newcommand{\U}{{\cal U}}

\newcommand{\resp}{resp.\ }

\newcommand{\ol}{\setlength{\itemsep}{0pt}\begin{enumerate}}
\newcommand{\eol}{\end{enumerate}\setlength{\itemsep}{-\parsep}}
\newcommand{\ul}{\setlength{\itemsep}{0pt}\begin{itemize}}
\newcommand{\dl}{\setlength{\itemsep}{0pt}\begin{description}}
\newcommand{\edl}{\end{description}\setlength{\itemsep}{-\parsep}}
\newcommand{\eul}{\end{itemize}\setlength{\itemsep}{-\parsep}}

\newcommand{\commentout}[1]{}

\newcommand{\bi}{\begin{itemize}}
\newcommand{\ei}{\end{itemize}}
\newcommand{\be}{\begin{enumerate}}
\newcommand{\ee}{\end{enumerate}}

\newcommand{\denselist}{\itemsep 0pt\partopsep 0pt}

\newenvironment{oldthm}[1]{\par\noindent{\bf Theorem #1:} \em \noindent}{\par}
\newenvironment{oldlem}[1]{\par\noindent{\bf Lemma #1:} \em \noindent}{\par}
\newenvironment{oldcor}[1]{\par\noindent{\bf Corollary #1:} \em \noindent}{\par}
\newenvironment{oldpro}[1]{\par\noindent{\bf Proposition #1:} \em \noindent}{\par}
\newcommand{\othm}[1]{\begin{oldthm}{\ref{#1}}}
\newcommand{\eothm}{\end{oldthm} \medskip}
\newcommand{\olem}[1]{\begin{oldlem}{\ref{#1}}}
\newcommand{\eolem}{\end{oldlem} \medskip}
\newcommand{\ocor}[1]{\begin{oldcor}{\ref{#1}}}
\newcommand{\eocor}{\end{oldcor} \medskip}
\newcommand{\opro}[1]{\begin{oldpro}{\ref{#1}}}
\newcommand{\eopro}{\end{oldpro} \medskip}

\newcommand{\bxor}[1]{\dot{\bor}}

\newcommand{\intension}[1]{[\![ #1 ]\!]}
\newcommand{\sK}{{\sf K}}
\newcommand{\sA}{{\sf A}}

\newcommand{\LKn}{{\cal L}^K_n}
\newcommand{\LKIn}{{\cal L}^{K,\hra}_n}
\newcommand{\LI}{{\cal L}^{\hra}}
\newcommand{\LKInt}{{\cal L}^{K,\hra}_n}
\newcommand{\LKAn}{{\cal L}^{K,X,A}_n}
\newcommand{\LXAn}{{\cal L}^{X,A}_n}
\newcommand{\LXAo}{{\cal L}^{X,A}_1}
\newcommand{\LXn}{{\cal L}^{X}_n}

\newcommand{\LKone}{{\cal L}^K_1}
\newcommand{\hra}{\hookrightarrow}
\newcommand{\dhra}{\rightleftharpoons}
\newcommand{\defined}{\downarrow\!\!}
\newcommand{\undefined}{\uparrow\!\!}

\newcommand{\AX}{{\rm AX}}

\newcommand{\AXKhran}{{\mathrm AX}^{K,\hra}_n}

\renewcommand{\H}{{\cal H}}

\newtheorem{FACT}[THEOREM]{Fact}
                            {\end{FACT}}

\begin{document}

\title{Interactive Unawareness Revisited\footnote{A preliminary
version of this paper was presented at the Tenth Conference on
Theoretical Aspects of Rationality and Knowledge (TARK05).}}

\author{Joseph Y. Halpern
\\ Computer Science Department \\ Cornell University,
U.S.A. \\  e-mail: halpern@cs.cornell.edu \and Leandro Chaves
R\^ego
\\ School of Electrical and Computer Engineering \\
Cornell University, U.S.A. \\
e-mail: lcr26@cornell.edu}

\date{}
\maketitle 

\begin{abstract}
We analyze a model of interactive unawareness introduced by
Heifetz, Meier and Schipper (HMS). We consider two axiomatizations
for their model, which capture different notions of validity.
These axiomatizations allow us to compare the HMS approach to both
the standard (S5) epistemic logic and two other approaches to
unawareness: that of Fagin and Halpern and that of Modica and
Rustichini.  We show that the differences between the HMS approach
and the others are mainly due to the notion of validity used and
the fact that the HMS is based on a 3-valued propositional logic.
\end{abstract}

\section{Introduction}
\label{intro}

Reasoning about knowledge has played a significant role in work in
philosophy, economics, and distributed computing.  Most of that
work has used standard Kripke structures to model knowledge, where
an agent knows a fact $\phi$ if $\phi$ is true in all the worlds
that the agent considers possible.  While this approach has proved
useful for many applications, it suffers from a serious
shortcoming, known as the {\em logical omniscience\/} problem
(first observed and named by Hintikka \citeyear{Hi1}): agents know
all tautologies and know all the logical consequences of their
knowledge.  This seems inappropriate for resource-bounded agents
and agents who are unaware of various concepts (and thus do not
know logical tautologies involving those concepts). To take just
one simple example, a novice investor may not be aware of the
notion of the price-earnings ratio, although that may be relevant
to the decision of buying a stock.

There has been a great deal of work on the logical omniscience
problem (see \cite{FHMV} for an overview). Of most relevance to
this paper are approaches that have focused on (lack of)
awareness.  Fagin and Halpern \citeyear{FH} (FH from now on) were
the first to deal with lack of model omniscience explicitly in
terms of awareness.  They did so by introducing an explicit
awareness operator.  Since then, there has been a stream of papers
on the topic in the economics literature (see, for example,
\cite{MR94,MR99,DLR98}). In these papers, awareness is defined in
terms of knowledge:~an agent is aware of $p$ if he either knows
$p$ or knows that he does not know $p$. All of them focused on the
single-agent case. Recently, Heifetz, Meier, and Schipper
\citeyear{HMS03} (HMS from now on) have provided a multi-agent
model for unawareness.  In this paper, we consider how the HMS
model compares to other work.

A key feature of the HMS approach (also present in the work of
Modica and Rustichini \citeyear{MR99}---MR from now on) is that
with each world or state is associated a (propositional) language.
Intuitively, this is the language of concepts defined at that
world.  Agents may not be aware of all these concepts.  The way
that is modeled is that in all the states an agent considers
possible at a state $s$, fewer concepts may be defined than are
defined at state $s$.  Because a proposition $p$ may be undefined
at a given state $s$, the underlying logic in HMS is best viewed
as a 3-valued logic: a proposition $p$ may be true, false, or
undefined at a given state.

We consider two sound and complete axiomatizations for the HMS
model, that differ with respect to the language used and the
notion of validity. One axiomatization captures {\em weak
validity\/}: a formula is weakly valid if it is never false
(although it may be undefined). In the single-agent case, this
axiomatization is identical to that given by MR.  However, in the
MR model, validity is taken with respect to ``objective'' state,
where all formulas are defined.  As shown by Halpern
\citeyear{Hal34}, this axiomatization is also sound and complete
in the single-agent case with respect to a special case of FH's
awareness structures; we extend Halpern's result to the
multi-agent case. The other axiomatization of the HMS model
captures {\em (strong) validity}: a formula is (strongly) valid if
it is always true. If we add an axiom saying that there is no
third value to this axiom system, then we just get the standard
axiom system for S5.  This shows that, when it comes to strong
validity,  the only difference between the HMS models and standard
epistemic models is the third truth value.

The rest of this paper is organized as follows. In
Section~\ref{backsection}, we review the basic S5 model, the FH
model, the MR model, and the HMS model. In
Section~\ref{sec:axHMS}, we compare the HMS approach and the FH
approach, both semantically and axiomatically, much as
Halpern~\citeyear{Hal34} compares the MR and FH approaches.  We
show that weak validity in HMS structures corresponds in a precise
sense to validity in awareness structures.  In
Section~\ref{sec:strongvalidity}, we extend the HMS language by
adding a nonstandard implication operator. Doing so allows us to
provide an axiomatization for strong validity. We conclude in
Section~\ref{sec:conclusion}. Further discussion of the original
HMS framework and an axiomatization of strong validity in the
purely propositional case can be found in the appendix.

\section{Background}
\label{backsection}

We briefly review the standard epistemic logic and the approaches of FH,
MR, and HMS here.

\subsection{Standard epistemic logic}
\label{S5section}
The syntax of standard epistemic logic is straightforward.
Given a set $\{1, \ldots, n\}$ of agents, formulas are formed by
starting with a set $\Phi = \{p, q, \ldots\}$ of primitive propositions
as well as a special formula $\top$ (which is always true), and then
closing off under conjunction ($\land$), negation ($\neg$)
and the modal operators $K_i$, $i = 1, \ldots, n$.  Call the
resulting language $\LKn(\Phi)$.%
\footnote{In MR, only the single-agent case is considered.  We
consider the multi-agent here to allow the generalization to HMS.
In many cases, $\top$ is defined in terms of other formulas, e.g.,
as $\neg (p \land \neg p)$.  We take it to be primitive here for
convenience.} As usual, we define $\phi \lor \psi$ and $\phi \rimp
\psi$ as abbreviations of $\neg (\neg \phi \land \neg \psi)$ and
$\neg \phi \lor \psi$, respectively.

The standard approach to giving semantics to $\LKn(\Phi)$ uses
Kripke structures. A {\it Kripke structure for $n$ agents (over
$\Phi$)} is a tuple $M= (\Sigma,\pi, {\cal K}_1, \ldots, \K_n)$,
where $\Sigma$ is a set of states, $\pi: \Sigma \times \Phi
\rightarrow \{0,1\}$ is an interpretation, which associates with
each primitive propositions its truth value at each state in
$\Sigma$, ${\cal K}_i: \Sigma \rightarrow 2^\Sigma$ is a {\it
possibility correspondence} for agent $i$. Intuitively, if $t\in
{\cal K}_i(s)$, then agent $i$ considers state $t$ possible at state
$s$. ${\cal K}_i$ is {\it reflexive} if for all $s\in \Sigma$, $s\in
{\cal K}_i(s)$; it is {\it transitive} if for all $s,t\in \Sigma$,
if $t\in{\cal K}_i(s)$ then $\K_i(t) \subseteq \K_i(s)$; it is {\it
Euclidean\/} if for all $s,t\in \Sigma$, if $t\in{\cal K}_i(s)$
then $\K_i(t) \supseteq \K_i(s)$.%
\footnote{It is more standard in the philosophy literature to take
$\K_i$ to be a binary relation.  The two approaches are
equivalent, since if $\K_i'$ is a binary relation, we can define a
possibility correspondence $\K_i$ by taking $t \in \K_i(s)$ iff
$(s,t) \in \K_i'$. We can similarly define a binary relation given
a possibility correspondence.  Given this equivalence, it is easy
to see that the notions of a possibility
correspondence being reflexive, transitive, or Euclidean are
equivalent to the corresponding notion for binary relations.} A
Kripke structure is reflexive (resp., reflexive and transitive;
partitional) if the possibility correspondences ${\cal K}_i$ are
reflexive (resp., reflexive and transitive; reflexive, Euclidean,
and transitive). Let $\M_n(\Phi)$ denote the class of all Kripke
structures for $n$ agents over $\Phi$, with no restrictions on the
$\K_i$ relations. We use the superscripts $r$, $e$, and $t$ to
indicate that the $\K_i$ relations are restricted to being
reflexive, Euclidean, and transitive, respectively.  Thus, for
example, $\M_n^{rt}(\Phi)$ is the class of all reflexive and
transitive Kripke structures for $n$ agents.

We write $(M,s)\sat \varphi$ if $\varphi$ is true at state $s$ in
the Kripke structure $M$. The truth relation is defined
inductively as follows:
\begin{eqnarray}
& & (M,s)\sat p\mbox{, for } p\in\Phi\mbox{, if }\pi(s,p)= 1
\nonumber \\
& & (M,s)\sat \neg \varphi \mbox{ if }(M,s)\not\sat\varphi
\nonumber \\
& & (M,s)\sat \varphi\wedge \psi  \mbox{ if }(M,s)\sat
\varphi\mbox{ and }(M,s)\sat \psi \nonumber\\
& & (M,s)\sat K_i \varphi \mbox{ if }(M,s')\sat \varphi\ \mbox{for
all }s'\in {\cal K}_i(s). \nonumber
\end{eqnarray}

A formula $\varphi$ is said to be {\it valid} in Kripke structure
$M$ if $(M,s)\sat\varphi$ for all $s\in \Sigma$. A formula
$\varphi$ is valid in a class $\N$ of Kripke structures, denoted
$\N\sat\varphi$, if it is valid for all Kripke structures in
${\cal N}$.

An {\em axiom system\/}\index{axiom system}~AX consists of a
collection of {\em axioms\/}\index{axiom|(} and {\em inference
rules}\index{inference rule|(}. An axiom is a formula, and an
inference rule has the form ``from $\phi_1, \ldots, \phi_k$
infer~$\psi$,'' where $\phi_1, \ldots, \phi_k, \psi$ are formulas.
A formula $\phi$ is {\it provable} in AX, denoted AX $\vdash
\phi$, if there is a sequence of formulas such that the last one
is $\phi$, and each one is either an axiom or follows from
previous formulas in the sequence by an application of an
inference  rule. An axiom system AX is said to be {\em sound} for
a language ${\cal L}$ with respect to a class ${\cal N}$ of
structures if every formula provable in AX is valid with respect
to ${\cal N}$. The system AX is {\em complete} for ${\cal L}$ with
respect to ${\cal N}$ if every formula in ${\cal L}$ that is valid
with respect to ${\cal N}$ is provable in AX.

Consider the following set of well-known axioms and inference
rules:

\begin{description}
\item[{\rm Prop.}] All substitution instances of valid formulas of
propositional logic.

\item[{\rm K.}] $(K_i\varphi\land K_i(\varphi\rimp\psi))\rimp
K_i\psi$.

\item[{\rm T.}] $K_i\varphi\rimp \varphi$.

\item[{\rm 4.}] $K_i\varphi\rimp K_iK_i\varphi$.

\item[{\rm 5.}] $\neg K_i\varphi\rimp K_i\neg K_i\varphi$.

\item[{\rm MP.}] {F}rom $\varphi$ and $\varphi\rimp\psi$ infer
$\psi$ (modus ponens).

\item[{\rm Gen.}] {F}rom $\varphi$ infer $K_i \varphi$.

\end{description}

It is well known that the axioms T, 4, and 5 correspond to the
requirements that the $\K_i$ relations are reflexive, transitive,
and Euclidean, respectively.  Let ${\bf K}_n$ be the axiom system
consisting of the axioms Prop, K and rules MP, and Gen, and let
${\bf S5}_n$ be the system consisting of all the axioms and
inference rules above. The following result is well known (see,
for example, \cite{Chellas,FHMV} for proofs).

\thm Let $\C$ be a (possibly empty) subset of $\{\rm{T}, 4, 5\}$
and let $C$ be the corresponding subset of $\{r, t, e\}$.  Then
${\bf K}_n \union \C$ is a sound and complete axiomatization of
the language $\LKn(\Phi)$ with respect to $\M_n^C(\Phi)$. \ethm In
particular, this shows that ${\bf S5}_n$ characterizes partitional
models, where the possibility correspondences are reflexive,
transitive, and Euclidean.

\subsection{The FH model}
\label{FHsection}

The Logic of General Awareness model of Fagin and Halpern
\citeyear{FH} introduces a syntactic notion of awareness. This is
reflected in the language by adding a new modal operator $A_i$ for
each agent $i$. The intended interpretation of $A_i\varphi$ is
``$i$ is aware of $\varphi$''. The power of this approach comes
from the flexibility of the notion of awareness. For example,
``agent $i$ is aware of $\varphi$'' may be interpreted as ``agent
$i$ is familiar with all primitive propositions in $\varphi$'' or
as ``agent $i$ can compute the truth value of $\varphi$ in time
$t$''.

Having awareness  in the language allows us to distinguish two
notions of knowledge: implicit knowledge and explicit knowledge.
Implicit knowledge, denoted with $K_i$, is defined as truth in all
worlds the agent considers possible, as usual. Explicit knowledge,
denoted with $X_i$, is defined as the conjunction of implicit
knowledge and awareness. Let $\LKAn(\Phi)$ be the language
extending $\LKn(\Phi)$ by closing off under the operators $A_i$
and $X_i$, for $i = 1, \ldots, n$. Let $\LXAn(\Phi)$ (\resp
$\LXn(\Phi)$) be the sublanguage of $\LKAn(\Phi)$ where the
formulas do not mention $K_1, \ldots, K_n$ (resp., $K_1, \ldots,
K_n$ and $A_1, \ldots A_n$).

An {\it awareness structure for $n$ agents over $\Phi$} is a tuple
$M =(\Sigma,\pi,{\cal K}_1,...,{\cal K}_n,{\cal A}_1,...,{\cal
A}_n)$, where $(\Sigma,\pi,{\cal K}_1,...,{\cal K}_n)$ is a Kripke
structure and ${\cal A}_i$ is a function associating a set of
formulas for each state, for $i= 1,...,n$. Intuitively, ${\cal
A}_i(s)$ is the set of formulas that agent $i$ is aware of at
state $s$. The set of formulas the agent is aware of can be
arbitrary. Depending on the interpretation of awareness one has in
mind, certain restrictions on ${\cal A}_i$ may apply. There are
two restrictions that are of particular interest here:
\begin{itemize}
\item Awareness is {\em generated by primitive propositions\/} if,
for all agents $i$, $\phi \in \A_i(s)$ iff all the primitive
propositions that appear in $\phi$ are in $\A_i(s) \inter \Phi$.
That is, an agent is aware of $\phi$ iff she is aware of all the
primitive propositions that appear in $\phi$. \item {\em Agents
know what they are aware of\/} if, for all
agents $i$, $t \in \K_i(s)$ implies that $\A_i(s) = \A_i(t)$.%
\end{itemize}
Following Halpern \citeyear{Hal34}, we say that awareness
structure is {\em propositionally determined\/} if awareness is
generated by primitive propositions and agents know what they are
aware of.

The semantics for awareness structures extends the semantics
defined for standard Kripke structures by adding two clauses
defining $A_i$ and $X_i$:
\begin{eqnarray}
\nonumber & & (M,s)\sat A_i\varphi\mbox{ if }\varphi\in {\cal
A}_i(s)\\
\nonumber & & (M,s)\sat X_i\varphi\mbox{ if }(M,s)\sat
A_i\varphi\mbox{ and }(M,s)\sat K_i\varphi.
\end{eqnarray}

FH provide a complete axiomatization for the logic of awareness;
we omit the details here.

\subsection{The MR model}
\label{MRsection}

We follow Halpern's \citeyear{Hal34} presentation of MR here; it
is easily seen to be equivalent to that in \cite{MR99}.

Since MR consider only the single-case, they use the language
$\LKone(\Phi)$. A {\em generalized standard model\/} (GSM) over
$\Phi$ has the form $M= (S,\Sigma,\pi,{\cal K},\rho)$, where
\begin{itemize}
\item $S$ and $\Sigma$ are disjoint sets of states; moreover,
$\Sigma= \cup_{\Psi\subseteq \Phi}S_{\Psi}$, where the sets
$S_{\Psi}$ are disjoint. Intuitively, the states in $S$ describe
the objective situation, while the states in $\Sigma$ describe the
agent's subjective view of the objective situation, limited to the
vocabulary that the agent is aware of. \item $\pi: S \times \Phi
\rimp \{0,1\}$ is an interpretation. \item ${\cal K}: S
\rightarrow 2^{\Sigma}$ is a {\em generalized possibility
correspondence}. \item $\rho$ is a projection from $S$ to $\Sigma$
such that (1) if $\rho(s)= \rho(t)\in S_{\Psi}$ then (a) $s$ and
$t$ agree on the truth values of all primitive propositions in
$\Psi$, that is, $\pi(s,p)= \pi(t,p)$ for all $p\in \Psi$ and (b)
${\cal K}(s)= {\cal K}(t)$ and (2) if $\rho(s)\in S_{\Psi}$, then
${\cal K}(s)\subseteq S_{\Psi}$. Intuitively, $\rho(s)$ is the
agent's subjective state in objective state $s$.
\end{itemize}

We can extend ${\cal K}$ to a map (also denoted ${\cal K}$ for
convenience) defined on $S\cup \Sigma$ in the following way: if
$s'\in \Sigma$ and $\rho(s)= s'$, define ${\cal K}(s')= {\cal
K}(s)$. Condition 1(b) on $\rho$ guarantees that this extension is
well defined. A GSM is reflexive (resp., reflexive and transitive;
partitional) if ${\cal K}$ restricted to $\Sigma$ is reflexive
(resp., reflexive and transitive; reflexive, Euclidean and
transitive). Similarly, we can extend $\pi$ to a function (also
denoted $\pi$) defined on $S\cup \Sigma$: if $s' \in S_\Psi$, $p
\in \Psi$ and $\rho(s) = s'$, define $\pi(s',p)=\pi(s,p)$; and if
$s' \in S_\Psi$ and $p \notin \Psi$, define $\pi(s',p) = 1/2$.

With these extensions of $\K$ and $\pi$, the semantics for
formulas in GSMs is identical to that in standard Kripke
structures except for the negation, which is defined as follows:
\begin{eqnarray}
& & \mbox{if $s \in S$, then $(M,s)\sat \neg\varphi$ iff
$(M,s)\not\sat\varphi$}\nonumber \\
& & \mbox{if $s \in S_{\Psi}$, then $(M,s)\sat \neg\varphi$ iff
$(M,s)\not\sat\varphi$ and $\varphi\in\LKone(\Psi)$.} \nonumber
\end{eqnarray}
Note that for states in the ``objective'' state space $S$, the
logic is 2-valued; and every formula is either true or false.  On
the other hand, for states in the ``subjective'' state space
$\Sigma$ the logic is 3-valued. A formula may be neither true nor
false.  It is easy to check that if $s \in S_\Psi$, then every
formula in $\LKone(\Psi)$ is either true or false  at $s$, while
formulas not in $\LKone(\Psi)$ are neither true nor false.
Intuitively, an agent can assign truth values only to formulas
involving concepts he is aware of; at states in $S_\Psi$, the
agent is aware only of concepts expressed in the language
$\LKone(\Psi)$.

The intuition behind MR's notion of awareness is that an agent is
unaware of $\varphi$ if he does not know $\varphi$, does not know
he does not know it, and so on. Thus, an agent is aware of $\phi$
if he either knows $\phi$ or knows he does not know $\phi$, or
knows that he does not know that he does not know $\phi$, or
\ldots. MR show that under appropriate assumptions, this infinite
disjunction is equivalent to the first two disjuncts, so they
define $A \phi$ to be an abbreviation of $K \phi \lor K \neg K
\phi$.

Rather than considering validity, MR consider what we call here
{\em objective validity}: truth in all objective states (that is,
the states in $S$). Note that all classical (2-valued)
propositional tautologies are objectively valid in the MR setting.
MR provide a system ${\cal U}$ that is a sound and complete
axiomatization for objective validity with respect to partitional
GSM structures. The system $\U$ consists of the axioms Prop, T,
and 4, the inference rule MP, and the following additional axioms
and inference rules:

\begin{description}
\item[{\rm M.}] $K(\varphi\land\psi)\rimp K\varphi\land K\psi$.

\item[{\rm C.}] $K\varphi\land K\psi\rimp K(\varphi\land\psi)$.

\item[{\rm A.}] $A\varphi\dimp  A\neg\varphi$.

\item[{\rm AM.}] $A(\varphi\land\psi)\rimp A\varphi\land A\psi$.

\item[{\rm N.}] $K\top$.

\item[{\rm RE$_{sa}$.}] {F}rom $\varphi\dimp \psi$ infer
$K\varphi\dimp  K\psi$, where $\varphi$ and $\psi$ contain exactly
the same primitive propositions.
\end{description}

\thm {\rm \cite{MR99}} ${\cal U}$ is a complete and sound
axiomatization of objective validity for the language
$\LKone(\Phi)$ with respect to partitional GSMs over $\Phi$. \ethm

\subsection{The HMS model}\label{sec:HMS}
HMS define their approach semantically, without giving a logic. We
discuss their semantic approach in the appendix. To facilitate
comparison of HMS to the other approaches we have considered, we
define an appropriate logic. (In recent work done independently of
ours \cite{HMS05}, HMS also consider a logic based on their
approach, whose syntax and semantics is essentially identical to
that described here.)

Given a set $\Phi$ of primitive propositions, consider again the
language $\LKn(\Phi)$.   An {\em HMS structure for $n$ agents\/}
(over $\Phi$) is a tuple $M = (\Sigma,\K_1, \ldots, \K_n,\pi,
\{\rho_{\Psi',\Psi}: \Psi \subseteq \Psi' \subseteq \Phi\} )$,
where (as in MR), $\Sigma = \union_{\Psi \subseteq \Phi} S_\Psi$
is a set of states, $\pi: \Sigma \times \Phi \rightarrow
\{0,1,1/2\}$ is an interpretation such that for $s \in S_\Psi$,
$\pi(s,p) \ne 1/2$  iff $p \in \Psi$ (intuitively, all primitive
propositions in $\Psi$ are defined at states of $S_\Psi$), and
$\rho_{\Psi',\Psi}$ maps $S_{\Psi'}$ onto $S_{\Psi}$. Intuitively,
$\rho_{\Psi',\Psi}(s)$ is a description of the state $s \in
S_{\Psi'}$ in the less expressive vocabulary of $S_\Psi$.
Moreover, if $\Psi_1 \subseteq \Psi_2 \subseteq \Psi_3$, then
$\rho_{\Psi_3,\Psi_2} \circ \rho_{\Psi_2,\Psi_1} =
\rho_{\Psi_3,\Psi_1}$. Note that although both MR and HMS have
projection functions, they have slightly different intuitions
behind them. For MR, $\rho(s)$ is the subjective state (i.e., the
way the world looks to the agent) when the actual objective state
is $s$.  For HMS, there is no objective state;
$\rho_{\Psi',\Psi}(s)$ is the description of $s$ in the less
expressive vocabulary of $S_{\Psi}$.
For $B\subseteq S_{\Psi_2}$, let
$\rho_{\Psi_2,\Psi_1}(B)=\{\rho_{\Psi_2,\Psi_1}(s):s\in B\}$.
Finally, the $\sat$ relation in HMS structures is defined for
formulas in $\LKn(\Phi)$ in exactly the same way as it is in
subjective states of MR structures. Moreover, like MR, $A_i \phi$
is defined as an abbreviation of $K_i \phi \lor K_i \neg K_i
\phi$.

Note that the definition of $\sat$ does not use the functions
$\rho_{\Psi,\Psi'}$.  These functions are used only to impose some
coherence conditions on HMS structures. To describe these
conditions, we need a definition. Given $B \subseteq S_{\Psi}$,
let $B^\uparrow = \union_{\Psi' \supseteq \Psi}
\rho^{-1}_{\Psi',\Psi}(B)$.  Thus, we can think of $B^\uparrow$ as
the states in which $B$ can be expressed.
\begin{enumerate}
\item Confinedness: If $s \in S_{\Psi}$ then $\K_i(s)\subseteq
S_{\Psi'}$ for some $\Psi'\subseteq\Psi$.

\item Generalized reflexivity: $s \in \K_i(s)^\uparrow$ for all $s
\in \Sigma$.

\item Stationarity: $s'\in {\cal K}_i(s)$ implies
\begin{itemize}
\item[(a)] ${\cal K}_i(s')\subseteq {\cal K}_i(s)$; \item[(b)]
${\cal K}_i(s')\supseteq {\cal K}_i(s)$.
\end{itemize}

\item Projections preserve knowledge: If $\Psi_1 \subseteq \Psi_2
\subseteq \Psi_3$, $s \in S_{\Psi_3}$, and ${\cal
K}_{i}(s)\subseteq S_{\Psi_2}$, then $\rho_{\Psi_2,\Psi_1}({\cal
K}_i(s))={\cal K}_i(\rho_{\Psi_3,\Psi_1}(s))$.

\item Projections preserve ignorance: If $s\in S_{\Psi'}$ and
$\Psi \subseteq \Psi'$ then $({\cal K}_i(s))^{\uparrow}\subseteq
({\cal K}_i(\rho_{\Psi',\Psi}(s)))^{\uparrow}$.\footnote{HMS
explicitly assume that $\K_i(s) \ne \emptyset$ for all $s \in
\Sigma$, but since this follows from generalized reflexivity we do
not assume it explicitly. HMS also mention one other property,
which they call {\em projections preserve awareness}, but, as HMS
observe, it follows from the assumption that projections preserve
knowledge, so we do not consider it here.}
\end{enumerate}

We remark that HMS combined parts (a) and (b) of stationarity into
one statement (saying $\K_i(s) = \K_i(s')$). We split the
condition in this way to make it easier to capture axiomatically.
Roughly speaking, generalized reflexivity, part (a) of
stationarity, and part (b) of stationarity are analogues of the
assumptions in standard epistemic structures that the possibility
correspondences are reflexive, transitive, and Euclidean,
respectively. The remaining assumptions can be viewed as coherence
conditions.  See \cite{HMS03} for further discussion of these
conditions.

If $C$ is a subset of $\{r,t,e\}$, let
$\H_n^C(\Phi)$ denote the class of HMS structures over $\Phi$
satisfying confinedness, projections preserve knowledge,
projections preserve ignorance, and the subset of generalized
reflexivity, part (a) of stationarity, and part (b) of
stationarity corresponding to $C$.
Thus, for
example, $\H^{rt}_n(\Phi)$ is the class of HMS structures for $n$
agents over $\Phi$ that satisfy confinedness, projections preserve
knowledge, projections preserve ignorance, generalized
reflexivity, and part (a) of stationarity. HMS consider only
``partitional'' HMS structures, that is, structures in
$\H_n^{rte}(\Phi)$.  However, we can get more insight into HMS
structures by allowing the greater generality of considering
non-partitional structures.

\section{A Comparison of the Approaches}\label{sec:axHMS}

As a first step to comparing the MR, HMS, and FH approaches, we recall a
result proved by Halpern.

\lem\label{Adefinable} {\rm \cite[Lemma 2.1]{Hal34}} If $M$ is a
partitional awareness structures where awareness is generated by
primitive propositions, then
$$M \sat A_i \phi \dimp (X_i \phi \lor (\neg X_i \phi \land X_i \neg X_i
\phi)).$$
\elem

\noindent Halpern proves this lemma only for the single-agent case, but
the proof goes through without change for the multi-agent case.  Note
that this equivalence does not hold in general in non-partitional
structures.

Thus, if we restrict to partitional awareness structures where awareness
is generated by primitive propositions, we can define awareness just as
MR and HMS do.

Halpern \citeyear[Theorem 4.1]{Hal34} proves an even stronger
connection between the semantics of FH and MR, essentially showing
that partitional GSMs are in a sense equivalent to propositionally
determined awareness structures.  We prove a generalization of
this result here.

If $C$ is a subset of $\{r,t,e\}$, let $\N_n^{C,pd}(\Phi)$
and $\N_n^{C,pg}$ denote the set of propositionally determined awareness
structures  over $\Phi$ and the set of awareness structures over $\Phi$
where awareness is propositionally generated, respectively,
whose $\K_i$ relations satisfy the conditions in $C$.
Given
a formula $\phi \in \LKn(\Phi)$, let $\phi_X \in \LXn(\Phi)$ be
the formula that results by replacing all occurrences of $K_i$ in
$\phi$ by $X_i$.  Finally, let $\Phi_\phi$ be the set of primitive
propositions appearing in $\phi$.

\thm\label{thm:equivalent} Let $C$ be a subset of $\{r,t,e\}$.
\begin{itemize}
\item[(a)] If $M = (\Sigma,\K_1, \ldots, \K_n,\pi,
\{\rho_{\Psi',\Psi}: \Psi \subseteq \Psi' \subseteq \Phi\}) \in
\H_n^C(\Phi)$, then there exists an awareness structure $M' =
(\Sigma, \K_1', \ldots, \K_n', \pi', \A_1, \ldots, \A_n) \in
\N_n^{C,pg}(\Phi)$ such that, for all $\phi\in \LKn(\Phi)$, if $s \in
S_{\Psi}$ and $\Phi_\phi \subseteq \Psi$, then $(M,s) \sat \phi$ iff
$(M',s) \sat \phi_X$.
Moreover, if $C \inter \{t,e\} \ne \emptyset$, then we can take
$M' \in \N_n^{C,pd}$.

\item[(b)] If $M = (\Sigma, \K_1, \ldots, \K_n, \pi, \A_1, \ldots,
\A_n) \in \N_n^{C,pd}(\Phi)$, then there exists an HMS structure
$M' =  (\Sigma',\K_1', \ldots,$ $\K_n',\pi', \{\rho_{\Psi',\Psi}:
\Psi \subseteq \Psi' \subseteq \Phi\}) \in \H_n^C(\Phi)$ such that
$\Sigma' = \Sigma \times 2^{\Phi}$, $S_{\Psi} = \Sigma \times
\{\Psi\}$ for all $\Psi \subseteq \Phi$, and, for all
$\phi\in\LKn(\Phi)$, if $\Phi_\phi \subseteq \Psi$, then $(M,s) \sat
\phi_X$ iff $(M',(s,\Psi)) \sat \phi$.
If $\{t,e \} \inter C= \emptyset$, then the result holds even if
$M \in (\N_n^{C,pg}(\Phi)-\N_n^{C,pd}(\Phi))$.

\end{itemize}
\ethm

It follows immediately from Halpern's analogue of
Theorem~\ref{thm:equivalent} that $\phi$ is objectively valid in
GSMs iff $\phi_X$ is valid in propositionally determined
partitional awareness structures.  Thus, objective validity in
GSMs and validity in propositionally determined partitional
awareness structures are characterized by the same set of axioms.

We would like to get a similar result here.  However, if we define
validity in the usual way---that is, $\phi$ is valid iff $(M,s)
\sat \phi$ for all states $s$ and all HMS structures $M$---then it
is easy to see that there are no (non-trivial) valid HMS formulas.
Since the HMS logic is three-valued, besides what we will call
{\em strong validity\/} (truth in all states), we can consider
another standard notion of validity. A formula is {\em weakly
valid\/} iff it is not false at any state in any HMS structure
(that is, it is either true or undefined at every state in every
HMS structure). Put another way, $\phi$ is weakly valid if, at all
states where $\phi$ is defined, $\phi$ is true.

\cor\label{cor:weakvalidity}
If $C \subseteq \{r,t,e\}$ then
\begin{itemize}
\item[(a)] if $C \inter \{t,e\}=\emptyset$, then $\phi$ is weakly
valid in $\H_n^C(\Phi)$ iff $\phi_X$ is valid in
$\N_n^{C,pg}(\Phi)$;
\item[(b)] if $C \inter \{t,e\} \ne \emptyset$, then $\phi$ is
weakly valid in $\H_n^C(\Phi)$ iff $\phi_X$ is valid in
$\N_n^{C,pd}(\Phi)$.
\end{itemize}
\ecor

Halpern~\citeyear{Hal34} provides a sound and complete
axiomatizations for the language $\LXAo(\Phi)$ with respect to
$\N^{C,pd}(\Phi)$, where $C$ is either $\emptyset$, $\{r\}$,
$\{r,t\}$ and $\{r,e,t\}$.  It is straightforward to extend his
techniques to other subsets of $\{r,e,t\}$ and to arbitrary
numbers of agents.  However, these axioms involve combinations of
$X_i$ and $A_i$; for example, all the systems have an axiom of the
form $X \phi \land X(\phi \rimp \psi) \land A \psi \rimp X \psi$.
There seems to be no obvious axiomatization for $\LXn(\Phi)$ that
just involves axioms in the language $\LXn(\Phi)$ except for the
special case of partitional awareness structures, where $A_i$ is
definable in terms of $X_i$ (see Lemma~\ref{Adefinable}), although
this may simply be due to the fact that there are no interesting
axioms for this language.

Let S5$_n^X$ be the $n$-agent version of the axiom system S5$_X$
that Halpern proves is sound and complete for $\LXn(\Phi)$ with
respect to $\N_n^{ret,pd}(\Phi)$ (so that, for example, the axiom
$X \phi \land X(\phi \rimp \psi) \land A \psi \rimp X \psi$
becomes $X_i \phi \land X_i(\phi \rimp \psi) \land A_i \psi \rimp
X_i \psi$, where now we view $A_i \phi$ as an abbreviation for
$X_i \phi \lor X_i \neg X_i \phi$).  Let S5$_n^K$ be the result of
replacing all occurrences of $X_i$ in formulas in S5$_n^X$ by
$K_i$. Similarly, let $\U_n$ be the $n$-agent version of the axiom
system $\U$ together with the axiom $A_i K_j \phi \dimp A_i
\phi$,\footnote{The single-agent version of this axiom, $A K \phi
\dimp A \phi$, is provable in $\U$, so does not have to be given
separately.} and let $\U^X_n$ be the result of replacing all
instances of $K_i$ in the axioms of $\U_n$ by $X_i$. HMS have
shown that there is a sense in which a variant of $\U_n$ (which is
easily seen to be equivalent to $\U_n$) is a sound and complete
axiomatization for HMS structures \cite{HMS05}. Although this is
not the way they present it, their results actually show that
$\U_n$ is a sound and complete axiomatization of weak validity
with respect to $\H_n^{ret}(\Phi)$.

Thus, the following is immediate from Corollary~\ref{cor:weakvalidity}.

\cor\label{thm:HMSweakvalidity} $\U_n$ and S5$_n^K$ are both sound
and complete axiomatization of weak validity for the language
$\LKn(\Phi)$ with respect to $\H_n^{ret}(\Phi)$; $\U_n^X$ and
S5$_n^X$ are both sound and complete axiomatizations of validity
for the language $\LXn(\Phi)$ with respect to
$\N_n^{ret,pd}(\Phi)$. \ecor

We can provide a direct proof that $\U_n$ and S5$_n^K$ (resp.,
$\U_n^X$ and S5$_n^X$) are equivalent, without appealing to
Corollary~\ref{cor:weakvalidity}.  It is easy to check that all
the axioms of $\U_n^X$ are valid in $\N_n^{ret,pd}(\Phi)$ and all
the inference rules of $\U_n^X$ preserve validity.  From the
completeness of S5$_n^X$ proved by Halpern, it follows that
anything provable in $\U_n^X$ is provable in S5$_n^X$, and hence
that anything provable in $\U_n$ is provable in S5$_n^K$.
Similarly, it is easy to check that all the axioms of S5$_n^K$ are
weakly valid in $\H_n^{ret}(\Phi)$, and the inference rules
preserve validity.  Thus, from the results of HMS, it follows that
everything provable in S5$_n^K$ is provable in $\U_n$ (and hence
that everything provable in S5$_n^X$ is provable in $\U_n^X$).

These results show a tight connection between the various approaches.
$\U$ is a sound and complete axiomatization for objective validity in
partitional GSMs; $\U_n$
is a sound and complete axiomatization for weak validity in partitional
HMS structures; and $\U_n^X$ is a sound and complete axiomatization for
(the standard notion of) validity in partitional awareness structures
where awareness is generated by
primitive propositions and agents know which formulas they are aware of.

\section{Strong Validity}\label{sec:strongvalidity}

We say a formula is {\em (strongly) valid\/} in HMS structures if
it is true at every state in every HMS structure. We can get
further insight into HMS structures by considering strong
validity.  However, since no nontrivial formulas in $\LKn(\Phi)$
are valid in HMS structures, we must first extend the language. We
do so by adding a nonstandard implication operator $\hra$ to the
language.\footnote{We remark that a nonstandard implication
operator was also added to the logic used by Fagin, Halpern, and
Vardi \citeyear{FHV3} for exactly the same reason, although the
semantics of the operator here is different from there, since the
underlying logic is different.} Given an HMS structure $M$, define
$\intension{\phi}_M = \{s: (M,s) \sat \phi\}$; that is,
$\intension{\phi}_M$ is the set of states in $M$ where $\phi$ is
true. Roughly speaking, we want to define $\hra$ in such a way
that if $\intension{\phi}_M \subseteq \intension{\psi}_M$, then
$\phi \hra \psi$ is valid in $M$.  The one time when we do not
necessarily want this is  if ${\phi}_M = \emptyset$.  For example,
we definitely do not want $r \lor (p \land \neg p) \hra r \lor (q
\land \neg q)$ to be valid (since $r \lor (p \land \neg p)$ will
be true at a state where $r$ is true, $p$ is defined, and $q$ is
undefined, while $r \land (q \land \neg q)$ is undefined at such a
state).  Thus, it seems unreasonable to have $p \land \neg p \hra
q \land \neg q$ be valid, even though $\intension{p \land \neg
p}_M = \emptyset$.  If $\intension{\phi}_M = \emptyset$, we take
$\phi \hra \psi$ to be valid only if $\psi$ is at least as defined
as $\phi$.  Since the set of states where $\psi$ is defined in $M$
is $\intension{\psi}_M \union \intension{\neg \psi}_M$, this
condition becomes $\intension{\neg \phi}_M \subseteq
\intension{\psi}_M \union \intension{\neg \psi}_M$.

Let $\LKIn(\Phi)$ be the language that results by closing off
under $\hra$ in addition to $\neg$, $\land$, and $K_1, \ldots,
K_n$; let $\LI(\Phi)$ be the propositional fragment of the
language. We cannot use the MR definition of negation for
$\LKIn(\Phi)$, since $\phi \hra \psi$ may be defined even in
states where $\phi$ and $\psi$ are not defined.  (For example, $p
\hra p$ is true in all states, even in states where $p$ is not
defined.)  Thus, we must separately define the truth and falsity
of all formulas at all states, which we do as follows.  In the
definitions, we use $(M,s) \sat \undefined\phi$ as an abbreviation
of $(M,s) \not\sat \phi$ and $(M,s) \not \sat \neg \phi$; and
$(M,s) \sat \defined \phi$ as an abbreviation of $(M,s) \sat \phi$
or $(M,s) \sat \neg \phi$ (so $(M,s) \sat \undefined\phi$ iff
$\phi$ is neither true nor false at $s$, i.e.,  it is undefined at
$s$).
\begin{eqnarray}
& & (M,s)\sat\top \nonumber\\
& & (M,s)\not\sat\neg\top \nonumber\\
& & (M,s)\sat p \mbox{ if }\pi(s,p)= 1 \nonumber\\
& & (M,s)\sat \neg p \mbox{ if }\pi(s,p)= 0 \nonumber\\
& & (M,s)\sat \neg \neg \varphi \mbox{ if }(M,s)\sat \varphi \nonumber\\
& & (M,s)\sat \varphi\land\psi\mbox{ if }(M,s)\sat\varphi\mbox{
and }(M,s)\sat\psi \nonumber\\
& & (M,s)\sat \neg(\varphi\land\psi)\mbox{ if either }(M,s)\sat
\neg\varphi \land \psi\mbox{ or } (M,s)\sat \varphi \land
\neg\psi\mbox{ or }
(M,s)\sat \neg \varphi \land \neg\psi\nonumber\\
& & (M,s)\sat (\varphi\hra\psi)\mbox{ if either $(M,s) \sat \phi
\land \psi$ or $(M,s) \sat \undefined \phi$ or $(M,s) \sat \neg
\phi  \land
\defined \psi$}\nonumber\\
& & (M,s)\sat \neg(\varphi\hra\psi)\mbox{ if
$(M,s) \sat \phi \land \neg \psi$}\nonumber\\
& & (M,s)\sat K_i\varphi\mbox{ if }(M,s)\sat \defined \phi \mbox{
and
$(M,t)\sat\varphi$ for all $t\in{\cal K}_i(s)$}\nonumber \\
& & (M,s)\sat \neg K_i\varphi\mbox{ if }(M,s)\not\sat
K_i\varphi\mbox{ and }(M,s)\sat \defined \varphi.\nonumber
\nonumber
\end{eqnarray}
It is easy to check that this semantics agrees with the MR
semantics for formulas in $\LKn(\Phi)$. Moreover, the following
lemma follows by an easy induction on the structure of formulas.

\lem\label{lem:defined} If $\Psi \subseteq \Psi'$, every formula in
$\LKIn(\Psi)$ is defined at every state in $S_{\Psi'}$.
\elem

It is useful to define the following abbreviations:
\begin{itemize}
\item $\phi \dhra \psi$ is an abbreviation of $(\phi \hra \psi)
\land (\psi \hra \phi)$; \item $\varphi= 1$ is an abbreviation of
$\neg(\varphi\hra\neg\top)$; \item $\varphi= 0$ is an abbreviation
of $\neg(\neg\varphi\hra\neg\top)$; \item $\varphi= \frac{1}{2}$
is an abbreviation of
$(\varphi\hra\neg\top)\land(\neg\varphi\hra\neg\top)$.
\end{itemize}
Using  the formulas $\phi = 0$, $\phi=\frac{1}{2}$, and $\phi =
1$, we can reason directly about the truth value of formulas. This
will be useful in our axiomatization.

In our axiomatization of $\LKIn(\Phi)$ with respect to HMS
structures, just as in standard epistemic logic, we focus on
axioms that characterize properties of the $\K_i$ relation that
correspond to reflexivity, transitivity, and the Euclidean
property.

Consider the following axioms:
\begin{description}
\item[{\rm Prop$'$.}] All substitution instances of formulas valid
in $\LI(\Phi)$.

\item[{\rm K$'$.}] $K_i \varphi \land K_i(\varphi\hra\psi))\hra
K_i\psi$.

\item[{\rm T$'$.}] $K_i\varphi\hra
\varphi\lor\bigvee_{\{p:p\in\Phi_{\varphi}\}}K_i(p= 1/2)$.

\item[{\rm 4$'$.}] $K_i\varphi\hra K_iK_i\varphi$.

\item[{\rm 5$'$.}] $\neg K_i\neg K_i\varphi\hra (K_i\varphi)\lor
K_i(\varphi= 1/2)$.

\item[{\rm Conf1.}] $(\varphi= 1/2)\hra K_i(\varphi= 1/2)$ if
$\phi\in \LKn(\Phi)$.

\item[{\rm Conf2.}] $\neg K_i(\varphi= 1/2)\hra
K_i((\varphi\lor\neg\varphi)= 1)$.

\item[{\rm B1.}] $(K_i\varphi)= 1/2 \dhra \varphi= 1/2$.

\item[{\rm B2.}] $((\varphi=0\lor\varphi=1)\land K_i(\varphi= 1))\hra (K_i\varphi)= 1$.

\item[{\rm MP$'$.}] From $\phi$ and $\phi \hra \psi$ infer $\psi$.

\end{description}

A few comments regarding the axioms: Prop$'$, K$'$, T$'$, 4$'$,
5$'$, and MP$'$ are weakenings of the corresponding axioms and
inference rule for standard epistemic logic. All of them use $\hra$
rather than $\rimp$; in some cases further weakening is required. We
provide an axiomatic characterization of Prop$'$ in the appendix. A
key property of the axiomatization is that if we just add the axiom
$\phi \ne 1/2$ (saying that all formulas are defined), we get a
complete axiomatization of classical logic. T (with $\rimp$ replaced
by $\hra$) is sound in HMS systems satisfying generalized
reflexivity for formulas $\phi$ in $\LKn(\Phi)$.  But, for example,
$K_i(p = 1/2) \hra p=1/2$ is not valid; $p$ may be defined (i.e., be
either true or false) at a state $s$ and undefined at all states $s'
\in \K_i(s)$.  Note that axiom 5 is equivalent to its contrapositive
$\neg K_i \neg K_i \phi \rimp K_i\phi$.  This is not sound in its
full strength; for example, if $p$ is defined at $s$ but undefined
at the states in $\K_i(s)$, then $(M,s) \sat \neg K_i \neg K_i p
\land \neg K_i p$. Axioms Conf1 and Conf2, as the names suggest,
capture confinedness.  We can actually break confinedness into two
parts. If $s \in S_\Psi$, the first part says that each state $s'
\in \K_i(s)$ is in some set $S_{\Psi'}$ such that $\Psi' \subseteq
\Psi$.  In particular, that means that a formula in $\LKn(\Phi)$
that is undefined at $s$ must be undefined at each state in
$\K_i(s)$.  This is just what Conf1 says. Note that Conf1 does not
hold for arbitrary formulas; for example, if $p$ is defined and $q$
is undefined at $s$, and both are undefined at all states in
$\K_i(s)$, then $(M,s) \sat (p \hra q) = 1/2 \land \neg K_i((p \hra
q) =1/2)$. The second part of confinedness says that all states in
$\K_i(s)$ are in the same set $S_{\Psi'}$.   This is captured by
Conf2, since it says that if $\phi$ is defined at some state in
$\K_i(s)$, then it is defined at all states in $\K_i(s)$.
B1 and B2 are technical axioms that capture the semantics of
$K_i\phi$.\footnote{We remark that axiom B2 is slightly modified
from the preliminary version of the paper.}
\commentout{
A rule essentially identical to IR1 was used by Halpern~\citeyear{Hal34},
who conjectured that it should be unnecessary.  We also suspect that
it is unnecessary here. \footnote{We remark that IR1 was not
mentioned in the preliminary version of this paper and that axiom B2
is slightly modified from the preliminary version.}
}

Let $\AXKhran$ be the system consisting of Prop$'$, K$'$, B1, B2,
Conf1, Conf2, MP$'$, and Gen.

\thm\label{thm:main} Let $\C$ be a (possibly empty) subset of
$\{\mathrm{T}', 4', 5'\}$ and let $C$ be the corresponding subset
of $\{r, t, e\}$.  Then
$\AXKhran \union \C$ is a sound and
complete axiomatization of the language $\LKIn(\Phi)$ with
respect to $\H_n^C(\Phi)$. \ethm
%
%


Theorem~\ref{thm:main} also allows us to relate HMS structures to
standard epistemic structures. It is easy to check that if $\C$ is
a (possibly empty) subset of $\{\mathrm{T}', 4', 5'\}$ and $C$ is
the corresponding subset of $\{r,e,t\}$, all the axioms of
$\AXKhran \union \C$  are sound with respect to standard epistemic
structures $\M_n^C(\Phi)$.  Moreover, we get completeness by
adding the axiom $\phi \ne 1/2$, which says that all formulas are
either true or false. Thus, in a precise sense, HMS differs from
standard epistemic logic by allowing a third truth value.

\section{Conclusion}\label{sec:conclusion}
We have compared the HMS approach and the FH approach to modeling
unawareness.  Our results show that, as long as we restrict to the
language $\LKn(\Phi)$, the approaches are essentially equivalent;
we can translate from one to the other.  We are currently
investigating extending the logic of awareness by allowing
awareness of unawareness \cite{HR05b}, so that it would be
possible to say, for example, that there exists a fact that agent
1 is unaware of but agent 1 knows that agent 2 is aware of it.
This would be expressed by the formula $\exists p (\neg A_1 p
\land X_1 A_2 p)$.  Such reasoning seems critical to capture what
is going on in a number of games.  Moreover, it is not clear
whether it can be expressed in the HMS framework.

\paragraph{Acknowledgments:} We thank Aviad Heifetz, Martin Meier, and
Burkhard Schipper for useful discussions on awareness. This work
was supported in part by NSF under grants CTC-0208535 and
ITR-0325453, by ONR under grants N00014-00-1-03-41 and
N00014-01-10-511, and by the DoD Multidisciplinary University
Research Initiative (MURI) program administered by the ONR under
grant N00014-01-1-0795. The second author was also supported in
part by a scholarship from the Brazilian Government through the
Conselho Nacional de Desenvolvimento Cient\'ifico e Tecnol\'ogico
(CNPq).

\bibliographystyle{chicagor}
\bibliography{z,joe}

\appendix

\section{The Original HMS Approach}\label{sec:HMSold}
HMS describe their approach purely semantically, without giving a
logic. We review their approach here (making some  inessential
changes for ease of exposition). An {\em HMS frame for $n$
agents\/} is a tuple $(\Sigma, \K_1, \ldots,$ $ \K_n, (\Delta,
\preceq), $ $\{\rho_{\beta,\alpha}: \alpha,\beta \in \Delta,
\alpha \preceq \beta\})$, where:
\begin{itemize}
\item $\Delta$ is an arbitrary lattice, partially ordered by
$\preceq$; \item $\K_1, \ldots, \K_n$ are possibility
correspondences, one for each agent; \item $\Sigma$ is a disjoint
union of the form $\union_{\alpha\in\Delta} S_\alpha$; \item if
$\alpha \preceq \beta$, then $\rho_{\beta, \alpha} : S_\beta
\rightarrow S_\alpha$ is a surjection.
\end{itemize}
In the logic-based version of HMS given in Section~\ref{sec:HMS},
$\Delta$ consists of the subsets of $\Phi$, and $\Psi \preceq
\Psi'$ iff $\Psi \subseteq \Psi'$. Thus, the original HMS
definition can be viewed as a more abstract version of that given
in Section~\ref{sec:HMS}.

Given $B \subseteq S_\alpha$, let $B^\uparrow = \union_{\{\beta:\
\alpha \preceq \beta\}} \rho^{-1}_{\beta,\alpha}(B)$. We can think
of $B^\uparrow$ as the states in which $B$ can be expressed. HMS
focus on sets of the form $B^\uparrow$, which they take to be
events.

HMS assume that their frames satisfy the five conditions mentioned
in Section~\ref{sec:HMS}, restated in their more abstract setting.
The statements of generalized reflexivity and stationarity remain
the same. Confinedness, projections preserve knowledge, and
projections preserve ignorance are stated as follows:
\begin{itemize}
\item confinedness: if $s \in S_{\beta}$ then $\K_i(s)\subseteq
S_{\alpha}$ for some $\alpha\preceq\beta$;

\item projections preserve knowledge: if
$\alpha\preceq\beta\preceq\gamma$, $s\in S_{\gamma}$, and
$\K_i(s)\subseteq S_{\beta}$, then
$\rho_{\beta,\alpha}(\K_i(s))=\K_i(\rho_{\gamma,\alpha}(s))$;

\item projections preserve ignorance: if $s\in S_{\beta}$ and
$\alpha\preceq\beta$ then $(\K_i(s))^{\uparrow}\subseteq
(\K_i(\rho_{\beta,\alpha}(s)))^{\uparrow}$.
\end{itemize}

HMS start by considering the algebra consisting of all events of
the form $B^\uparrow$.  In this algebra, they define an operator
$\neg$ by taking $\neg(B^\uparrow) = (S_\alpha - B)^\uparrow$ for
$\emptyset \ne B\subseteq S_{\alpha}$. With this definition, $\neg
\neg B^\uparrow = B^\uparrow$ if $B \notin \{
\emptyset,S_\alpha\}$. However, it remains to define $\neg
\emptyset^\uparrow$.  We could just take it to be $\Sigma$, but
then we have $\neg \neg S_\alpha^\uparrow = \Sigma$, rather than
$\neg \neg S_\alpha^\uparrow = S_\alpha^\uparrow$. To avoid this
problem, in their words, HMS ``devise a distinct vacuous event
$\emptyset^{S_\alpha}$'' for each subspace $S_\alpha$, extend the
algebra with these events, and define $\neg S_\alpha^\uparrow =
\emptyset^{S_\alpha}$ and $\neg \emptyset^{S_\alpha} =
S_\alpha^{\uparrow}$. They do not make clear exactly what it means
to ``devise a vacuous event''. We can recast their definitions in
the following way, that allows us to bring in the events
$\emptyset^{S_\alpha}$ more naturally.

In a 2-valued logic, given a formula $\phi$ and a structure $M$,
the set $\intension{\phi}_M$ of states where $\phi$ is true and
the set $\intension{\neg \phi}_M$ of states where $\phi$ is false
are complements of each other, so it suffices to associate with
$\phi$ only one set, say $\intension{\phi}_M$.  In a 3-valued
logic, the set of states where $\phi$ is true does not determine
the set of states where $\phi$ is false. Rather, we must consider
three mutually exclusive and exhaustive sets: the set where
$\varphi$ is true, the set where $\varphi$ is false, and the set
where $\varphi$ is undefined. As before, one of these is
redundant, since it is the complement of the union of the other
two. Note that if $\varphi$ is a formula in the language of HMS,
the set $\intension{\phi}_M$ is either $\emptyset$ or an event of
the form $B^{\uparrow}$, where $B\subseteq S_{\alpha}$.  In the
latter case, we associate with $\phi$ the pair of sets
$(B^{\uparrow}, (S_\alpha - B)^{\uparrow})$, i.e.,
$(\intension{\phi}_M, \intension{\neg \phi}_M)$. In the former
case, we must have $\intension{\neg \phi}_M = S_\alpha^\uparrow$
for some $\alpha$, and we associate with $\phi$ the pair
$(\emptyset, S_\alpha^\uparrow)$. Thus, we are using the pair
$(\emptyset, S_\alpha^\uparrow)$ instead of devising a new event
$\emptyset^{S_\alpha}$ to represent $\intension{\phi}_M$ in this
case.\footnote{In a more recent version of their paper, HMS
identify a nonempty event $E$ with the pair $(E,S)$, where, for
$E=B^{\uparrow}$, $S$ is the unique set $S_\alpha$ containing $B$.
Then $\emptyset^S$ can be identified with $(\emptyset,S)$.  While
we also identify events with pairs of sets and $\emptyset^S$ with
$(\emptyset,S)$, our identification is different from that of HMS,
and extends more naturally to sets that are not events.}

HMS use intersection of events to represent conjunction.  It is
not hard to see that the intersection of events is itself an
event.  The obvious way to represent disjunction is as the union
of events, but the union of events is in general not an event.
Thus, HMS define a disjunction operator using de Morgan's law: $E
\lor E' = \neg(\neg E \inter \neg E')$.  In our setting, where we
use pairs of sets, we can also define operators $\sim$ and
$\sqcap$ (intuitively, for negation and intersection) by taking
$\sim\!(E,E') = (E',E)$ and
$$(E,E') \sqcap (F,F') = (E \inter F, (E \inter F') \union (E'
\inter F) \union (E' \inter F')).$$
Although our definition of $\sqcap$ may not seem so intuitive, as the
next result shows, $(E,E') \sqcap (F,F')$ is essentially
equal to $(E \inter F, \neg (E \inter F))$.  Moreover, our definition
has the advantage of not using $\neg$, so it applies even if $E$ and $F$
are not events.

\lem \label{lem:unionofpair} If $(E\union
E')=S_{\alpha}^{\uparrow}$ and $(F\union
F')=S_{\beta}^{\uparrow}$, then
$$
(E \inter F') \union (E' \inter F) \union (E' \inter F') = \left\{
\begin{array}{ll}
\neg (E \inter F) &\mbox{if $(E \inter F) \ne \emptyset$,}\\
S_\gamma^{\uparrow} &\mbox{if $(E \inter F) = \emptyset$ and
$\gamma = \sup(\alpha,\beta)$.\footnotemark}
\end{array}
\right.
$$
\footnotetext{Note that
$\sup(\alpha,\beta)$ is well defined since $\Delta$ is a lattice.}
\elem

\prf Let $I$ be the set $(E\inter F)\union(E\inter
F')\union(E'\inter F)\union(E'\inter F')$. We first show that
$I=S_{\gamma}^{\uparrow}$, where $\gamma = \sup(\alpha,\beta)$. By
assumption, $E=B^{\uparrow}$ for some $B\subseteq S_{\alpha}$, and
$F=C^{\uparrow}$ for some $C\subseteq S_{\beta}$. Suppose that $s\in
I$. We claim that $s\in S_{\delta}$, where $\alpha\preceq\delta$ and
$\beta\preceq\delta$. Suppose, by way of contradiction, that
$\alpha\not\preceq\delta$. Then $s\notin E\union E'$, so $s\notin
I$, a contradiction. A similar argument shows that
$\beta\preceq\delta$. Thus $\gamma\preceq\delta$ and $s\in
S_{\gamma}^{\uparrow}$. For the opposite inclusion, suppose that
$s\in S_{\gamma}^{\uparrow}$. Since $\alpha\preceq\gamma$ and
$\beta\preceq\gamma$, the projections $\rho_{\gamma,\alpha}(s)$ and
$\rho_{\gamma,\beta}(s)$ are well defined.
Let $X=E$ if $\rho_{\gamma,\alpha}(s)\in B$ and $X=E'$ otherwise.
Similarly, let $Y=F$
if $\rho_{\gamma,\beta}(s)\in C$ and $Y=F'$ otherwise. It is easy to
see that $s\in (X\inter Y)\subseteq I$. It follows that $(E\inter
F')\union(E'\inter F)\union(E'\inter
F')=S_{\gamma}^{\uparrow}-(E\inter F)$. The result now follows
easily. \eprf

Finally, HMS define an operator $\sK_i$ corresponding to the
possibility correspondence $\K_i$.  They define
$\sK_i (E) = \{s:{\cal K}_i(s)\subseteq E\}$,%
\footnote{Actually, this is their definition only if $\{s: \K_i(s)
\subseteq E\} \ne \emptyset$; otherwise, they take $\sK_i(E) =
\emptyset^{S_\alpha}$ if $E=B^{\uparrow}$ for some $B \subseteq
S_\alpha$. We do not need a special definition if $\{s: \K_i(s)
\subseteq E\}=\emptyset$ using our approach.} and show that
$\sK_i(E)$ is an event if $E$ is.  In our setting, we define
$$\sK_i((E,E'))=
(\{s:{\cal K}_i(s)\subseteq E\}\cap(E\cup E'), (E \union E') -
\{s:{\cal K}_i(s)\subseteq E \}).$$ Essentially, we are defining
$\sK_i((E,E'))$ as $(\sK_i(E), \neg \sK_i(E))$.  Intersecting with
$E \union E'$ is unnecessary in the HMS framework, since their
conditions on frames guarantee that $\sK_i(E) \subseteq E \union
E'$. If we think of $(E,E')$ as $(\intension{\phi}_M,
\intension{\neg \phi}_M)$, then $\phi$ is defined on $E \union E'$.
The definitions above guarantee that $K_i\varphi$ is defined on the
same set.

HMS define an awareness operator in the spirit of MR, by taking
$\textsf{A}_i(E)$ to be an abbreviation of $\sK_i(E) \lor \sK_i
\neg \sK_i(E)$.  They then prove a number of properties of
knowledge and awareness, such as $\sK_i(E) \subseteq
\sK_i\sK_i(E)$ and $\sA_i(\neg E) = \sA_i(E)$.

The semantics we have given for our logic matches that of the operators
defined by HMS, in the sense of the following lemma.
\lem\label{HMS->HR}
For all formulas $\varphi,\psi\in \LKIn(\Phi)$ and all HMS structures $M$.
\begin{enumerate}
\item[(a)] $ (\intension{\neg\varphi}_M,
\intension{\neg\neg\phi}_M) = \, \sim\!(\intension{\varphi}_M,
\intension{\neg\phi}_M))$

\item[(b)] $(\intension{\varphi\land\psi}_M, \intension{\neg
(\varphi\land\psi)}_M) = (\intension{\varphi}_M, \intension{\neg
\varphi}_M) \sqcap (\intension{\psi}_M, \intension{\neg \psi}_M)$.
\item[(c)] $(\intension{K_i\varphi}_M,\intension{\neg K_i
\phi}_M)= \sK_i((\intension{\varphi}_M,\intension{\neg
\varphi}_M))$
\end{enumerate}
\elem

\prf Part (a) follows easily from the fact that $\,
\sim\!(\intension{\varphi}_M,
\intension{\neg\phi}_M))=(\intension{\neg\phi}_M),
\intension{\varphi}_M) $ and
$\intension{\neg\neg\phi}_M=\intension{\varphi}_M$.

For part (b), note that
$$(\intension{\varphi}_M,
\intension{\neg \varphi}_M)\sqcap (\intension{\psi}_M,
\intension{\neg
\psi}_M)=(\intension{\varphi}_M\inter\intension{\psi}_M,(\intension{\varphi}_M\inter\intension{\neg
\psi}_M)\union(\intension{\neg
\varphi}_M\inter\intension{\psi}_M)\union(\intension{\neg
\varphi}_M\inter\intension{\neg \psi}_M)).$$
Now the result is immediate from the observation that
$\intension{\varphi}_M\inter\intension{\psi}_M=\intension{\varphi\land\psi}_M$
and
$$(\intension{\varphi}_M\inter\intension{\neg
\psi}_M)\union(\intension{\neg
\varphi}_M\inter\intension{\psi}_M)\union(\intension{\neg
\varphi}_M\inter\intension{\neg \psi}_M)=\intension{\neg
(\varphi\land\psi)}_M.$$

For (c), by definition of $\sK_i$,
$$\sK_i((\intension{\varphi}_M,\intension{\neg\varphi}_M))=
(\{s:{\cal K}_i(s)\subseteq
\intension{\varphi}_M\}\cap(\intension{\varphi}_M\cup
\intension{\neg\varphi}_M), (\intension{\varphi}_M \union
\intension{\neg\varphi}_M) - \{s:{\cal K}_i(s)\subseteq
\intension{\varphi}_M\}).$$ Note that $t\in
(\intension{\varphi}_M\cup \intension{\neg\varphi}_M)$ iff
$(M,t)\sat\defined \varphi$, and $t\in\{s:{\cal K}_i(s)\subseteq
\intension{\varphi}_M\}$ iff for all $t'\in {\cal K}_i(t)$,
$(M,t')\sat\varphi$. Thus, $t\in \{s:{\cal K}_i(s)\subseteq
\intension{\varphi}_M\}\inter (\intension{\varphi}_M\cup
\intension{\neg\varphi}_M)$ iff $(M,t)\sat K_i\varphi$, i.e., iff
$t\in \intension{K_i\varphi}_M$. Similarly, $t\in
(\intension{\varphi}_M \union \intension{\neg\varphi}_M) - \{s:{\cal
K}_i(s)\subseteq \intension{\varphi}_M\}$ iff $(M,t)\sat\defined
\varphi$ and $(M,t)\not\sat K_i\varphi$, i.e., iff $(M,t)\sat\neg
K_i\varphi$. Hence, $t\in \intension{\neg K_i\varphi}_M$. \eprf

Note that Lemma~\ref{HMS->HR} applies even though, once we introduce
the $\hra$ operator, $\intension{\phi}_M$ is not in general an event
in the HMS sense. (For example, $\intension{p \hra q}_M$ is not in
general an event.)

\section{An Axiomatization of $\LI(\Phi)$}
Note that the formulas $\phi = 0$, $\phi=\frac{1}{2}$, and $\phi =
1$ are 2-valued.  More generally, we define a formula $\phi$ to be
{\em 2-valued\/} if $(\phi=0) \lor (\phi=1)$ is valid in all HMS
structures. Because they obey the usual axioms of classical logic,
2-valued formulas play a key role in our axiomatization of
$\LI(\Phi)$. We say that a formula is {\em definitely 2-valued\/}
if it is in the smallest set containing $\top$ and all formulas of
the form $\phi = k$ which is closed under negation, conjunction,
nonstandard implication, and $K_i$, so that if $\phi$ and $\psi$
are definitely two-valued, then so are $\neg \phi$, $\phi \land
\psi$, $\phi' \hra \psi$ (for all $\phi'$), and $K_i \phi$. Let $D_2$
denote the set
of definitely 2-valued formulas.

The following lemma is easy to prove. \lem If $\phi$ is definitely
2-valued, then it is 2-valued. \elem

Let AX$_3$ consist of the following collection of axioms and
inference rules:
\begin{enumerate}
\denselist

\item[P0.] $\top$.

\item[P1.] $(\varphi\land \psi) \dhra \neg(\varphi\hra\neg \psi)$
if $\phi, \psi\in D_2$.

\item[P2.] $\varphi\hra (\psi\hra \varphi)$ if $\phi, \psi\in
D_2$.

\item[P3.] $(\varphi\hra (\psi\hra \phi'))\hra((\varphi\hra
\psi)\hra(\varphi\hra \phi'))$ if $\varphi, \psi$, $\phi'\in D_2$.

\item[P4.] $(\varphi\hra \psi)\hra ((\varphi\hra\neg \psi)\hra\neg
\varphi)$ if $\phi, \psi\in D_2$.

\item[P5.] $(\varphi\land\psi)= 1\dhra(\varphi= 1)\land(\psi= 1)$.

\item[P6.] $(\varphi\land\psi)= 0\dhra(\varphi= 0\land\neg(\psi=
1/2))\lor(\neg(\varphi= 1/2)\land\psi= 0)$.

\item[P7.] $\varphi= 1\dhra(\neg\varphi)= 0$.

\item[P8.] $\varphi= 0\dhra(\neg\varphi)= 1$.

\item[P9.] $(\varphi\hra\psi)= 1\dhra((\varphi= 0\land\neg(\psi=
1/2))\lor (\varphi=1/2) \lor (\varphi= 1\land\psi= 1))$.

\item[P10.] $(\varphi\hra\psi)= 0\dhra(\varphi= 1\land \psi=0)$.

\item[P11.]
$(\varphi= 0\lor\varphi= 1/2\lor\varphi= 1)\land(\neg(\varphi= i\land
\varphi= j))$,
for $i,j\in\{0,1/2,1\}$ and $i\neq j$.

\item[R1.] From $\phi=1$ infer $\phi$.

\item[MP$'$.] {F}rom $\phi$ and $\phi \hra \psi$ infer $\psi$.
\end{enumerate}

It is well
known that P0-P4 together with MP$'$ provide a complete
axiomatization for classical 2-valued propositional logic with
negation, conjunction, implication, and $\top$.\footnote{We remark
that we included formulas of the form $K_i \phi$ among the formulas
that are definitely 2-valued.  While such formulas are not relevant
in the axiomatization of $\LI(\Phi)$, they do play a role when we
consider the axiom Prop$'$ in $\AXKhran$, which applies to instances
in the language $\LKIn(\Phi)$ of valid formulas of $\LI(\Phi)$.}
Axioms P5-P10 are basically a translation to formulas of the
semantics for conjunction, negation and implication.

Note that all the axioms of AX$_3$ are sound in classical logic
(all formulas of the form $\phi = 1/2$ are vacuously false in
classical logic).  Moreover, it is easy to show that if we add the
axiom $\neg(\varphi= 1/2)$ to AX$_3$, we get a sound and complete
axiomatization of classical propositional logic (although many
axioms then become redundant).

\thm \label{axiomprop} AX$_3$ is a sound and complete
axiomatization of $\LI(\Phi)$. \ethm

\prf The proof that the axiomatization is sound is a straightforward
induction on the length of the proof of any theorem $\varphi$. We
omit the details here. For completeness, we need to show that a
valid formula $\phi \in \LI(\Phi)$ is provable in AX$_3$. We first
prove that $\phi=1$ is provable in AX$_3$ using standard techniques,
and then apply R1 to infer $\phi$. We proceed as follows.

Given a set ${\cal G}$ of formulas, let $\land{\cal G}=
\bigwedge_{\phi\in{\cal G}}\phi$. A set ${\cal G}$ of formulas is
{\em AX-consistent}, if for all finite subsets ${\cal
G}'\subseteq{\cal
G}$, $\AX\not\vdash\neg(\land{\cal G}')$. A set ${\cal G}$ of
formulas is {\em maximal AX-consistent} if ${\cal G}$ is
AX-consistent and for all $\varphi\notin{\cal G}$, ${\cal
G}\cup\{\varphi\}$ is not AX-consistent.

\lem \label{maximlemma} If ${\cal G}$ is an AX-consistent subset of
${\cal G'}$, then ${\cal G}$ can be extended to a maximal
AX-consistent subset of ${\cal G'}$. \elem

\prf The proof uses standard techniques. Let $\psi_1,\psi_2,\ldots$ be an
enumeration of the formulas in ${\cal G'}$. Define $F_0= {\cal G}$
and $F_i= F_{i-1}\cup\{\psi_{i}\}$ if $F_{i-1}\cup\{\psi_{i}\}$ is
AX-consistent and $F_i= F_{i-1}$, otherwise. Let ${\cal F}= \cup_{i=
0}^{\infty}F_i$. We claim that ${\cal F}$ is an maximal
AX-consistent subset of ${\cal G'}$. Suppose that $\psi\in{\cal G'}$
and $\psi\notin{\cal F}$. By construction, we have $\psi= \psi_k$
for some $k$. If $F_{k-1}\cup\{\psi_k\}$ were AX-consistent, then
$\psi_k$ would be in $F_k$ and hence $\psi_k$ would be in ${\cal
F}$. Since $\psi_k= \psi\notin{\cal F}$, we have that
$F_{k-1}\cup\{\psi\}$ is not AX-consistent and hence ${\cal
F}\cup\{\psi\}$, is not AX-consistent. \eprf

The next lemma shows that maximal $\AX_3$-consistent sets of
definitely 2-valued formulas satisfy essentially the same properties
as maximal classically consistent sets of formulas.

\lem \label{properlemma} Let AX be any axiom system that includes
AX$_3$. For all maximal AX-consistent subsets ${\cal F}$ of $D_2$,
the following properties hold:

\begin{enumerate}
\item[(1)] for every formula $\varphi\in D_2$, exactly one of
$\varphi$ and $\neg\varphi$ is in ${\cal F}$;

\item[(2)] for every formula $\varphi\in\LI(\Phi)$, exactly one of
$\varphi= 0$, $\varphi= 1/2$, and $\varphi= 1$ is in ${\cal F}$;

\item[(3)] if $\phi_1, \ldots, \phi_k,\psi\in D_2$, $\phi_1, \ldots,
\phi_k\in {\cal F}$, and $\AX_3\vdash(\phi_1 \land \ldots \land
\phi_k) \hra\psi$, then $\psi\in{\cal F}$;

\item[(4)] $(\varphi\land\psi)= 1\in{\cal F}$ iff $\varphi= 1\in
{\cal F}$ and $\psi= 1\in {\cal F}$;

\item[(5)] $(\varphi\land\psi)= 0\in{\cal F}$ iff either $\varphi=
0\in {\cal F}$ and $\psi= 1/2\notin{\cal F}$, or $\psi= 0\in {\cal
F}$ and $\varphi= 1/2\notin{\cal F}$;

\item[(6)] $\psi= 1\in{\cal F}$ iff $(\neg\psi)= 0\in{\cal F}$;

\item[(7)] $\psi= 0\in{\cal F}$ iff $(\neg\psi)= 1\in{\cal F}$;


\item[(8)] $(\varphi\hra\psi)= 1\in{\cal F}$ iff either $\varphi=
0\in{\cal F}$ and $\psi= 1/2\notin{\cal F}$; or $\varphi=
1/2\in{\cal F}$; or $\varphi= 1\in{\cal F}$ and $\psi= 1\in{\cal
F}$;

\item[(9)] $(\varphi\hra\psi)= 0\in{\cal F}$ iff $\varphi=
1\in{\cal F}$ and $\psi= 0\in{\cal F}$;

\item[(10)] if $\varphi\in D_2$ and $\AX\vdash\varphi$, then
$\varphi\in{\cal F}$;

\end{enumerate}
\elem

\prf First, note that axioms P0-P4 and MP$'$ guarantee that
classical propositional reasoning can be used for formulas in
$D_2$. We thus use classical propositional reasoning with minimal
comment.

For (1), we first show that exactly one of ${\cal F}\cup\{\varphi\}$
and ${\cal F}\cup\{\neg \varphi\}$ is AX-consistent. Suppose that
${\cal F}\cup\{\varphi\}$ and ${\cal F}\cup\{\neg\varphi\}$ are both
AX-consistent. Then $\varphi\in{\cal F}$ and $\neg\varphi\in{\cal
F}$. Since $\neg(\varphi\land\neg\varphi)\in D_2$,
$\AX\vdash\neg(\varphi\land\neg\varphi)$, but this is a
contradiction since ${\cal F}$ is AX-consistent. Now suppose that
neither ${\cal F}\cup\{\varphi\}$ nor ${\cal F}\cup\{\neg\varphi\}$
is AX-consistent. Then there exist finite subsets $H_1,H_2\subseteq
{\cal F}$ such that $\AX\vdash\neg(\varphi\land(\land H_1))$ and
$\AX\vdash\neg(\neg\varphi\land(\land H_2))$. Let $G= H_1\cup H_2$.
By classical propositional reasoning,
$\AX\vdash\neg(\varphi\land(\land G))$ and
$\AX\vdash\neg(\neg\varphi\land(\land G))$, so
$\AX\vdash\neg((\varphi\land(\land G))\lor(\neg\varphi\land(\land
G)))$ and $\AX\vdash\neg((\varphi\land(\land
G))\lor(\neg\varphi\land(\land G)))\hra\neg(\land G)$. Hence, by
MP$'$, $\AX\vdash\neg(\land G)$. This is a contradiction, since
$G\subseteq {\cal F}$ and ${\cal F}$ is AX-consistent.

Suppose that ${\cal F}\cup\{\varphi \}$ is AX-consistent (the other
case is completely analogous). Since ${\cal F}$ is a maximal
AX-consistent subset of $D_2$ and $\varphi\in D_2$, we have
$\varphi\in{\cal F}$. And since ${\cal F}\union\{\neg\varphi\}$ is
not AX-consistent, $\neg\varphi\notin{\cal F}$.

For (2), we first show that exactly one of ${\cal F}\cup\{\varphi=
i\}$, for $i= \{0,1/2,1\}$, is AX-consistent. Suppose that ${\cal
F}\cup\{\varphi= i\}$ and ${\cal F}\cup\{\varphi= j\}$, $i\neq j$,
are AX-consistent.
Then $\varphi= i\in{\cal F}$ and $\varphi= j\in{\cal F}$. By axiom
P11, $\AX\vdash\neg(\varphi= i\land\varphi= j)$. This is a
contradiction, since ${\cal F}$ is AX-consistent.

Next, suppose that none of ${\cal F}\cup\{\varphi= i\}$ is
AX-consistent. Then there exist finite sets $F_i\subseteq {\cal F}$
such that $\AX\vdash\neg(\varphi= i\land(\land F_i))$, $i=0,1/2,1$.
Let $G= F_0\cup F_{1/2}\cup F_1$. By classical propositional
reasoning, $\AX\vdash\neg(\varphi= i\land(\land G))$, and
$\AX\vdash\neg((\varphi= 0\land(\land G))\lor(\varphi=
1/2\land(\land G))\lor(\varphi= 1\land(\land G)))$. Now using axiom
P11, we have $\AX\vdash\neg(\land G)$. This is a contradiction,
since $G\subseteq {\cal F}$ and ${\cal F}$ is AX-consistent.

Let $i^{*}$ be the unique $i$ such that ${\cal F}\cup\{\varphi=
i^{*}\}$ is AX-consistent. Since ${\cal F}$ is a maximal
AX-consistent subset of $D_2$ and $\varphi= i^{*}\in D_2$, we have
that $\{\varphi= i^{*}\}\in{\cal F}$. And since ${\cal F}$ is
AX-consistent, it is clear by P11, that if $j\neq i^{*}$, then
$\{\varphi= j\}\notin{\cal F}$.

For (3), by part (1), if $\psi\notin{\cal F}$, then
$\neg\psi\in{\cal F}$. Thus,  $\{\phi_1, \ldots,
\phi_k,\neg\psi\}\subseteq{\cal F}$.   But since $\AX_3\vdash\phi_1
\land \ldots \land \phi_k \hra\psi$, by classical propositional
reasoning,
$\AX_3\vdash\neg(\phi_1\land\ldots\land\phi_k\land\neg\psi)$, a
contradiction since ${\cal F}$ is AX-consistent.

The proof of the remaining properties follows easily from parts (2)
and (3).  For example, for part (4), if $(\phi \land \psi) = 1 \in \F$,
then the fact that $\phi = 1 \in \F$ and $\psi = 1 \in F$ follows from
P5 and (3).  We leave details to the reader.
\eprf

A formula $\varphi$ is said to be {\it satisfiable in a structure
$M$} if $(M,s)\sat\varphi$ for some world in $M$; $\varphi$ is {\it
satisfiable in a class of structures ${\cal N}$} if it is
satisfiable in at least one structure in ${\cal N}$. Let ${\cal
M}_P$ be the class of all 3-valued propositional HMS models.

\lem \label{lemconsat} If $\varphi= i$ is AX$_3$-consistent, then
$\varphi= i$ is satisfiable in ${\cal M}_P$, for $i\in\{0,1/2,1\}$.
\elem

\prf We construct a special model $M^{c}\in{\cal M}_{P}$ called the
{\em canonical 3-valued model}. $M^{c}$ has a state $s_{V}$
corresponding to every $V$ that is a maximal AX$_3$-consistent
subset of $D_2$. We show that

$$(M^{c},s_V)\sat\varphi= j\mbox{\ \ iff\ \ }\varphi= j\in V\mbox{,
for }j\in\{0,1/2,1\}.$$

Note that this claim suffices to prove Lemma \ref{lemconsat} since,
by Lemma \ref{maximlemma}, if $\varphi= i$ is AX$_3$-consistent,
then it is contained in a maximal AX$_3$-consistent subset of $D_2$.
We proceed as follows. Let $M^{c}= (\Sigma,\pi)$, where $\Sigma=
\{s_V:V\mbox{ is a maximal consistent subset of $D_2$}\}$ and
\[
\pi(s_V,p)= \left\{
\begin{array}{ll}
                                1 & \mbox{if $p= 1\in V$,} \\
                                0 & \mbox{if $p= 0\in V$,} \\
                                1/2 & \mbox{if $p= 1/2\in V$.}\\
                                \end{array}
                            \right.
                            \]
Note that by Lemma \ref{properlemma}(2), the interpretation $\pi$
is well defined.

We now show that the claim holds by induction on the structure of
formulas. If $\psi$ is a primitive proposition, this follows from
the definition of $\pi(s_V,\psi)$.

Suppose that $\psi= \neg\phi$. By Lemma \ref{properlemma}(7),
$(\neg\phi)= 1\in V$ iff $\phi= 0\in V$. By the induction
hypothesis, $\phi= 0\in V$ iff $(M^{c},s_V)\sat \phi=0$. By the
semantics of the logic, we have $(M^{c},s_V)\sat \phi=0$ iff
$(M^{c},s_V)\sat\neg\phi$, and the latter holds iff
$(M^{c},s_V)\sat(\neg\phi)= 1$. Similarly, using Lemma
\ref{properlemma}(6), we can show $(\neg\phi)=0\in V$ iff
$(M^{c},s_V)\sat(\neg\phi)= 0$. The remaining case $\varphi=1/2$
follows from the previous cases, axiom P11, and the fact that
$(M^{c},s_V)\sat\psi= i$ for exactly one $i\in\{0,1/2,1\}$. (For all
the following steps of the induction the case $\varphi=1/2$ is
omitted since it follows from the other cases for exactly the same
reason.)

Suppose that $\psi= \phi_1\land\phi_2$. By Lemma
\ref{properlemma}(4), $\psi= 1\in V$ iff $\phi_1= 1\in V$ and
$\phi_2= 1\in V$. By the induction hypothesis, $\phi_j= 1\in V$ iff
$(M^{c},s_V)\sat\phi_j=1$ for $j\in{1,2}$, which is true iff
$(M^{c},s_V)\sat(\phi_1\land\phi_2)=1$. Similarly, using Lemma
\ref{properlemma}(5), we can show that $(\phi_1\land\phi_2)=0\in V$
iff $(M^{c},s_V)\sat(\phi_1\land\phi_2)= 0$.

Suppose that $\psi= \phi_1\hra\phi_2$. By Lemma
\ref{properlemma}(8), $\psi= 1\in V$ iff either $\phi_1= 0\in V$ and
$\phi_2= 1/2\notin V$; or $\phi_1= 1/2\in V$; or $\phi_1= 1\in V$
and $\phi_2= 1\in V$. By the induction hypothesis, this is true iff
either $(M^{c},s_V)\sat\phi_1=0$ and $(M^{c},s_V)\not\sat
\phi_2=1/2$; or $(M^{c},s_V)\sat\phi_1=1/2$; or
$(M^{c},s_V)\sat\phi_1=1$ and $(M^{c},s_V)\sat\phi_2=1$.
This, in turn,
is true iff $(M^{c},s_V)\sat\neg\phi_1$ and $(M^{c},s_V)\sat
\neg(\phi_2=1/2)$; or $(M^{c},s_V)\sat(\phi_1=1/2)$; or
$(M^{c},s_V)\sat\phi_1$ and $(M^{c},s_V)\sat\phi_2$. By the
semantics of $\hra$, this holds iff
$(M^{c},s_V)\sat(\phi\hra\psi)= 1$. Similarly, using Lemma
\ref{properlemma}(9), we can show that $(\phi\hra\psi)=0\in V$ iff
$(M^{c},s_V)\sat(\phi\hra\psi)= 0$. \eprf

We can finally complete the proof of Theorem \ref{axiomprop}.
Suppose that $\varphi$ is valid. This implies $(\varphi=
0)\lor(\varphi=1/2)$ is not satisfiable. By Lemma \ref{lemconsat},
$(\varphi= 0)\lor(\varphi=1/2)$ is not AX$_3$-consistent, so
$\AX_3\vdash\neg((\varphi= 0)\lor(\varphi=1/2))$. By axioms P0-P4,
P11 and MP$'$, $\AX_3\vdash\varphi= 1$. And finally,
applying R1, $\AX_3\vdash\varphi$. So, the axiomatization is
complete. \eprf


\section{Proofs of Theorems}

In this section, we provide proofs
of the theorems in Sections~\ref{sec:axHMS}
and~\ref{sec:strongvalidity}.  We restate the results for the
reader's convenience.

The next lemma, which is easily proved by induction on the structure
of formulas, will be used throughout.  We leave the proof to the
reader.

\lem \label{lem:moredefined}
If $M\in H_n(\Phi)$, $\Psi'\subseteq\Psi\subseteq\Phi$, $s\in\Psi$,
$s'=\rho_{\Psi,\Psi'}(s)$, $\varphi\in\LKn(\Phi)$, and
$(M,s')\sat\varphi$, then $(M,s)\sat\varphi$. \elem

\commentout{
\subsection{Proof of Theorem \ref{thm:HMSweakvalidity}}

HMS showed that $\U_n(\Phi)$ is a sound and complete
axiomatization if validity defined with respect to the states in
$S_{\Phi}$. We claim this is equivalent to $\U_n(\Phi)$ being
sound and complete with respect to weakly validity. To verify
soundness is straightforward; we omit the details. For
completeness suppose that $\phi$ is weakly valid, since all
formulas in $\LKn(\Phi)$ have truth value either 0 or 1 in all
states in $S_{\Phi}$, $\phi$ must have truth value 1 in all states
in $S_{\Phi}$, so it is valid in the HMS sense. Therefore, by the
completeness result of HMS, $\phi$ is a theorem of $\U_n(\Phi)$,
as desired. \bbox }

\bigskip

\othm{thm:equivalent} Let $C$ be a subset of $\{r,t,e\}$.
\begin{itemize}
\item[(a)] If $M = (\Sigma,\K_1, \ldots, \K_n,\pi,
\{\rho_{\Psi',\Psi}: \Psi \subseteq \Psi' \subseteq \Phi\}) \in
\H_n^C(\Phi)$, then there exists an awareness structure $M' =
(\Sigma, \K_1', \ldots, \K_n', \pi', \A_1, \ldots, \A_n) \in
\N_n^{C,pg}(\Phi)$ such that, for all $\phi\in\LKn(\Phi)$, if $s \in
S_{\Psi}$ and $\Phi_\phi \subseteq \Psi$, then $(M,s) \sat \phi$ iff
$(M',s) \sat \phi_X$.
Moreover, if $C \inter \{t,e\} \ne \emptyset$, then we can take
$M' \in \N_n^{C,pd}$.

\item[(b)] If $M = (\Sigma, \K_1, \ldots, \K_n, \pi, \A_1, \ldots,
\A_n) \in \N_n^{C,pd}(\Phi)$, then there exists an HMS structure
$M' =  (\Sigma',\K_1', \ldots,$ $\K_n',\pi', \{\rho_{\Psi',\Psi}:
\Psi \subseteq \Psi' \subseteq \Phi\}) \in \H_n^C(\Phi)$ such that
$\Sigma' = \Sigma \times 2^{\Phi}$, $S_{\Psi} = \Sigma \times
\{\Psi\}$ for all $\Psi \subseteq \Phi$, and, for all
$\phi\in\LKn(\Phi)$, if $\Phi_\phi \subseteq \Psi$, then $(M,s) \sat
\phi_X$ iff $(M',(s,\Psi)) \sat \phi$.
If $\{t,e \} \inter C= \emptyset$, then the result holds even if
$M \in (\N_n^{C,pg}(\Phi)-\N_n^{C,pd}(\Phi))$.
\end{itemize}
\eothm

\prf
For part (a), given $M = (\Sigma,\K_1, \ldots, \K_n,\pi,
\{\rho_{\Psi',\Psi}: \Psi \subseteq \Psi' \subseteq \Phi\}) \in
\H_n^C(\Phi)$, let $M' = (\Sigma,\K_1', \ldots,\K_n',\pi', \A_1,
\ldots,\A_n)$ be an awareness structure such that
\begin{itemize}
\item
$\pi'(s,p)=\pi(s,p)$ if $\pi(s,p)\ne 1/2$ (the definition of $\pi'$
if $\pi(s,p)=1/2$ is irrelevant);
\item
$\K'_i(s)=\K_i(s)$ if $\K_i$ does not satisfy
Generalized Reflexivity, and $\K'_i(s)=\K_i(s)\union \{s\}$
otherwise;
\item if $\emptyset\ne\K_i(s)\subseteq S_{\Psi}$ or if $\K_i(s)=\emptyset$ and $s\in
S_{\Psi}$, then $\A_i(s)$ is the smallest set of formulas containing
$\Psi$ that is
propositionally generated.
\end{itemize}
By construction, $M' \in \N_n^{C,pg}(\Phi)$. It is easy to check
that if $C \inter \{t,e\} \ne \emptyset$, then agents know what they
are aware of, so that $M' \in \N_n^{C,pd}(\Phi)$.

We complete the proof of part (a) by proving, by induction on the
structure of $\phi$, that if $s \in S_\Psi$ and $\Phi_\phi \subseteq
\Psi$, then $(M,s) \sat \phi$ iff $(M',s) \sat \phi_X$. If $\varphi$
is either a primitive proposition, or $\varphi=\neg\psi$, or
$\varphi=\varphi_1\land\varphi_2$, the result is obvious either from
the definition of $\pi'$ or from the induction hypothesis. We omit
details here.

Suppose that $\varphi=K_i\psi$. If $(M,s)\sat K_i\psi$, then for all
$t\in\K_i(s)$, $(M,t)\sat\psi$. By the induction hypothesis, it
follows that for all $t\in\K_i(s)$, $(M',t)\sat\psi$. If $\K_i$
satisfies generalized reflexivity, it easily follows from Lemma
\ref{lem:moredefined} that $(M,s)\sat\psi$ so, by the induction
hypothesis, $(M',s)\sat\psi$. Hence, for all $t\in\K'_i(s)$,
$(M',t)\sat\psi$, so $(M',s)\sat K_i\psi$.
To show that $(M',s) \sat X_i \psi$, it remains to show that $(M',s)
\sat A_i \psi$, that is, that $\psi \in \A_i(s)$.
First suppose that $\emptyset\ne\K_i(s)\subseteq S_\Psi$.
Since $(M,t)\sat\psi$ for all $t\in \K_i(s)$, it follows
that $\psi$ is defined at all states in $\K_i(s)$. Thus,
$\Phi_{\psi}\subseteq\Psi$, for otherwise a simple induction shows that
$\psi$ would be undefined at states in $S_{\Psi}$. Hence,
$\Phi_{\psi}\subseteq\A_i(s)$.
Since awareness is generated by primitive propositions, we have $\psi
\in \A_i(s)$, as desired.
Now suppose that $\K_i(s)=\emptyset$ and $s\in S_\Psi$.  By
assumption, $\Phi_\phi=\Phi_\psi\subseteq\Psi\subseteq\A_i(s)$,
so again $\psi \in \A_i(s)$.

For the converse, if $(M's)\sat X_i\psi$, then $(M',s)\sat K_i\psi$
and $\psi\in \A_i(s)$. By the definition of $\A_i$,
$\K_i(s)\subseteq S_{\Psi}$, where $\Phi_\psi\subseteq\Psi$. Since
$(M',s)\sat K_i\psi$, $(M',t)\sat\psi$ for all $t\in\K'_i(s)$.
Therefore, by the induction hypothesis, $(M,t)\sat\psi$ for all
$t\in\K_i(s)$, which implies $(M,s)\sat K_i\psi$, as desired.

\commentout{ Consider $\varphi=A_i\psi$. Since
$A_i\psi=K_i\psi\lor K_i\neg K_i\psi$ in the HMS model, the fact
that if $(M,s)\sat\varphi$ then $(M',s)\sat\varphi$ follows from
the previous cases. For the converse, if $(M',s)\sat A_i\psi$,
then $\psi$ contains only primitive propositions that are defined
at all states in $\K_i(s)$. Hence, $(M,s)\sat
K_i(\psi\lor\neg\psi)$. Then, to prove that $(M,s)\sat A_i\psi$,
it suffices to show that $(M,s)\sat K_i(\psi\lor\neg\psi)\rimp
K_i\psi\lor K_i\neg K_i\psi$. As $(M,s)\sat
K_i(\psi\lor\neg\psi)$, it follows that $\psi$ is defined at all
states $t\in\K_i(s)$. Then, there are two cases to consider: (1)
$(M,t)\sat \psi$ for all $t\in\K_i(s)$, (2) there exists
$t\in\K_i(s)$ such that $(M,t)\not\sat \psi$. For case (1), it
follows that $(M,s)\sat K_i\psi$, hence $(M,s)\sat K_i\psi\lor
K_i\neg K_i\psi$. For case (2), as $\K_i$ satisfies at $e$, for
any $t\in\K_i(s)$ we have $\K_i(t)\subseteq\K_i(s)$. Therefore,
$\psi$ is defined at all states in $\K_i(t)$ (since $\K_i(t)$ and
$\K_i(s)$ are in the same state space), which implies
$(M,t)\sat\neg K_i\psi$. Hence $(M,s)\sat K_i\neg K_i\psi$, so
$(M,s)\sat K_i\psi\lor K_i\neg K_i\psi$, as desired.}


For part (b), given $M = (\Sigma, \K_1, \ldots, \K_n, \pi, \A_1,
\ldots, \A_n) \in \N_n^{C,pd}(\Phi)$, let $M' =  (\Sigma',\K_1',
\ldots,$ $\K_n',\pi',
\\ \mbox{$\{\rho_{\Psi',\Psi}: \Psi \subseteq \Psi'
\subseteq \Phi\}$})$ be an HMS structure such that

\begin{itemize}
\item $\Sigma' = \Sigma \times 2^{\Phi}$;

\item $S_{\Psi} = \Sigma \times \{\Psi\}$ for $\Psi \subseteq
\Phi$;

\item $\pi'((s,\Psi),p)=\pi(s,p)$ if $p\in\Psi$ and
$\pi'((s,\Psi),p)=1/2$ otherwise;

\item $\K'_i((s,\Psi))=\{(t,\Psi\inter\Psi_i(s)):t\in\K_i(s)\}$, where
$\Psi_i(s)=\{p:p\in\A_i(s)\}$ is the set of primitive
propositions that agent $i$ is aware of at state $s$;

\item $\rho_{\Psi',\Psi}((s,\Psi'))=(s,\Psi)$.
\end{itemize}
Note that since agents know what they are aware of, if
$t\in\K_i(s)$, then $\Psi_i(t)=\Psi_i(s)$.

We first show that $M'$ satisfies confinedness and that projections preserve
knowledge and ignorance. Confinedness follows since
$\K'_i((s,\Psi))\subseteq S_{\Psi\inter\Psi_i(s)}$. To prove
projections preserve knowledge, suppose that
$\Psi_1\subseteq\Psi_2\subseteq\Psi_3$ and
$\K_i((s,\Psi_3))\subseteq S_{\Psi_2}$. Then
$\Psi_2=\Psi_3\inter\Psi_i(s)$ and $\Psi_1=\Psi_1\inter\Psi_i(s)$.
Thus
$\rho_{\Psi_2,\Psi_1}(\K'_i((s,\Psi_3)))=\{(t,\Psi_1):(t,\Psi_3\inter\Psi_i(s))\in\K'_i((s,\Psi_3))\}=\{(t,\Psi_1):t\in\K_i(s)\}$.
Similarly,
$\K'_i(\rho_{\Psi_3,\Psi_1}(s,\Psi_3))=\{(t,\Psi_1\inter\Psi_i(s)):t\in\K_i(s)\}$.
Therefore, projections preserve knowledge.

To prove that projections preserve ignorance, note that
$\K'_i(\rho_{\Psi',\Psi}(s,\Psi'))=\K'_i((s,\Psi))=\{(t,\Psi\inter\Psi_i(s)):t\in
\K_i(s)\}$ and $\K'_i((s,\Psi'))=\{(t,\Psi'\inter\Psi_i(s)):t\in
\K_i(s)\}$. If $(s,\Psi'')\in(\K'_i((s,\Psi')))^{\uparrow}$, then
$\Psi'\inter\Psi_i(s)\subseteq\Psi''$. Since $\Psi\subseteq\Psi'$,
it follows that $\Psi\inter\Psi_i(s)\subseteq\Psi''$. Hence
$(s,\Psi'')\in(\K'_i(\rho_{\Psi',\Psi}(s,\Psi')))^{\uparrow}$.
Therefore, projection preserves ignorance.

We now show by induction on the structure of $\phi$ that if
$\Phi_\phi \subseteq \Psi$, then $(M,s) \sat \phi_X$ iff
$(M',(s,\Psi)) \sat \phi$. If $\varphi$ is a primitive proposition,
or $\varphi=\neg\psi$, or $\varphi=\varphi_1\land\varphi_2$, the
result is obvious either from the definition of $\pi'$ or from the
induction hypothesis. We omit details here.

Suppose that $\varphi=K_i\psi$. If $(M',(s,\Psi))\sat K_i\psi$, then
for all $(t,\Psi\inter\Psi_i(s))\in\K'_i((s,\Psi))$,
$(M',(t,\Psi\inter\Psi_i(s)))\sat\psi$. By the induction hypothesis
and the definition of $\K'_i$, it follows that for all
$t\in\K_i(s)$, $(M,t)\sat\psi$, so $(M,s)\sat K_i\psi$. Also note
that if $(M',(t,\Psi\inter\Psi_i(s)))\sat\psi$, then $\psi$ is
defined at all states in $S_{\Psi\inter\Psi_i(s)}$, and therefore at
all states in $S_{\Psi_i(s)}$. Hence,
$\Phi_{\psi}\subseteq\Psi_i(s)\subseteq\A_i(s)$. Since awareness is
generated by primitive propositions, $\psi\in\A_i(s)$. Thus,
$(M,s)\sat A_i\psi$, which implies that $(M,s)\sat X_i\psi$, as
desired.

For the converse, suppose that $(M,s)\sat X_i\psi$ and
$\Phi_{\phi}\subseteq \Psi$. Then $(M,s)\sat K_i\psi$ and $\psi\in
\A_i(s)$. Since $\psi\in \A_i(s)$, $\Phi_\psi\subseteq\Psi_i(s)$.
Hence, $\Phi_\psi\subseteq(\Psi\inter\Psi_i(s))$. $(M,s)\sat
K_i\psi$ implies that $(M,t)\sat\psi$ for all $t\in\K_i(s)$.  By the
induction hypothesis, since $\Phi_\psi\subseteq\Psi$,
$(M',(t,\Psi))\sat\psi$ for all $t\in\K_i(s)$. Since
$\Phi_\psi\subseteq(\Psi\inter\Psi_i(s))$, by Lemma
\ref{lem:moredefined}, it follows that
$(M',(t,\Psi\inter\Psi_i(s)))\sat\psi$ for all
$(t,\Psi\inter\Psi_i(s))\in\K'_i((s,\Psi))$. Thus,
$(M',(s,\Psi))\sat K_i\psi$, as desired.

\commentout{ Consider $\varphi=A_i\psi$. Since
$A_i\psi=K_i\varphi\lor K_i\neg K_i\varphi$ in the HMS model, the
fact that if $(M,(s,\Psi))\sat\varphi$ then $(M',s)\sat\varphi$
follows from the previous cases. If $(M',s)\sat A_i\psi$, then
$\psi$ contains only primitive propositions that are in
$\Psi_i(s)$. Hence, for any $\Psi$ containing all the primitive
propositions in $\psi$, $(M,(t,\Psi\inter\Psi_i(s)))\sat
\psi\lor\neg\psi$, which implies for all
$(t,\Psi\inter\Psi_i(s))\in\K_i(s,\Psi)$,
$(M,(t,\Psi\inter\Psi_i(s)))\sat \psi\lor\neg\psi$. So,
$(M,(s,\Psi))\sat K_i(\psi\lor\neg\psi)$, which implies
$(M,(s,\Psi))\sat A_i(\psi)$, as desired.}

We now show that $M'\in \H_n^C(\Phi)$. If $\K_i$ is reflexive, then
$(s,\Psi\inter\Psi_i(s))\in\K'_i((s,\Psi))$, so
$(s,\Psi)\in(\K'_i((s,\Psi)))^{\uparrow}$.
Thus,
$M'$ satisfies generalized reflexivity. Now suppose that $\K_i$ is
transitive. If $(s',\Psi\inter\Psi_i(s))\in\K'_i((s,\Psi))$, then
$s'\in\K_i(s)$ and
$\K'_i((s',\Psi\inter\Psi_i(s)))=\{(t,\Psi\inter\Psi_i(s)\inter\Psi_i(s')):t\in\K_i(s')\}$.
Since agents know what they are aware of, if $s'\in\K_i(s)$, then
$\Psi_i(s)=\Psi_i(s')$, so
$\K'_i((s',\Psi\inter\Psi_i(s)))=\{(t,\Psi\inter\Psi_i(s)):t\in\K_i(s')\}$.
Since $\K_i$ is transitive, $\K_i(s')\subseteq\K_i(s)$, so
$\K'_i((s',\Psi\inter\Psi_i(s)))\subseteq\K'_i((s,\Psi))$. Thus,
$M'$ satisfies part (a) of stationarity. Finally, suppose that
$\K_i$ is
Euclidean. If $(s',\Psi\inter\Psi_i(s))\in\K'_i((s,\Psi))$, then
$s'\in\K_i(s)$ and
$\K'_i((s',\Psi\inter\Psi_i(s)))=\{(t,\Psi\inter\Psi_i(s)\inter\Psi_i(s')):t\in\K_i(s')\}$.
Since agents know what they are aware of, if $s'\in\K_i(s)$, then
$\Psi_i(s)=\Psi_i(s')$, so
$\K'_i((s',\Psi\inter\Psi_i(s)))=\{(t,\Psi\inter\Psi_i(s)):t\in\K_i(s')\}$.
Since $\K_i$ is Euclidean, $\K_i(s')\supseteq\K_i(s)$, so
$\K'_i((s',\Psi\inter\Psi_i(s)))\supseteq\K'_i((s,\Psi))$. Thus $M'$
satisfies part (b) of stationarity.

If $\{t,e \} \inter C= \emptyset$, then it is easy to check that the
result holds even if $M \in (\N_n^{C,pg}(\Phi)-\N_n^{C,pd}(\Phi))$,
since the property that agents know what they are aware of was only
used to prove part (a) and part (b) of stationarity
in the proof. \eprf

\ocor{cor:weakvalidity}
If $C \subseteq \{r,t,e\}$ then
\begin{itemize}
\item[(a)] if $C \inter \{t,e\} = \emptyset$, then $\phi$ is weakly valid in
$\H_n^C(\Phi)$ iff $\phi_X$ is valid in $\N_n^{C,pg}(\Phi)$.
\item[(b)] if $C \inter \{t,e\} \ne \emptyset$, then $\phi$ is
weakly valid in $\H_n^C(\Phi)$ iff $\phi_X$ is valid in
$\N_n^{C,pd}(\Phi)$.
\end{itemize}
\eocor

\prf For part (a), suppose that $\varphi_X$ is valid with respect
to the class of awareness structures $\N_n^{C,pg}(\Phi)$ where
awareness is generated by primitive propositions and that
$\varphi$ is not weakly valid with respect to the class of HMS
structures $\H_n^{C}(\Phi)$. Then $\neg\varphi$ is true at some
state in some HMS structure in $\H_n^{C}(\Phi)$. By part (a) of
Theorem \ref{thm:equivalent}, $\neg\varphi_X$ is also true at some
state in some awareness structure where awareness is
 generated by primitive propositions, a contradiction since $\varphi_X$ is
valid in $\N_n^{C,pg}(\Phi)$.

For the converse, suppose that $\varphi$ is weakly valid in
$\H_n^{C}(\Phi)$ and that $\varphi_X$ is not valid with respect to
the class of awareness structures $\N_n^{C,pg}(\Phi)$. Then
$\neg\varphi_X$ is true at some state in some awareness structure
in $\N_n^{C,pg}(\Phi)$. By part (b) of Theorem
\ref{thm:equivalent}, $\neg\varphi$ is also true at some state in
some HMS structure in $\H_n^{C}(\Phi)$, a contradiction since
$\varphi$ is weakly valid in $\H_n^{C}(\Phi)$.

The proof of part (b) is the same except that $\N_n^{C,pg}(\Phi)$ is
replaced throughout by $\N_n^{C,pd}(\Phi)$. \eprf

\commentout{ For part (b), suppose that $\varphi_X$ is valid with
respect to the class of awareness structures $\N_n^{C,pd}(\Phi)$
where awareness is propositionally determined and that $\varphi$ is
not weakly valid with respect to the class of HMS structures
$\H_n^{C}(\Phi)$. Then $\neg\varphi$ is true at some state in some
HMS structure in $\H_n^{C}(\Phi)$. By part (a) of Theorem
\ref{thm:equivalent}, $\neg\varphi_X$ is also true at some state in
some awareness structure where awareness is propositionally
determined, a contradiction since $\varphi_X$ is valid in
$\N_n^{C,pd}(\Phi)$.

For the converse, suppose that $\varphi$ is weakly valid in
$\H_n^{C}(\Phi)$ and that $\varphi_X$ is not valid with respect to
the class of awareness structures $\N_n^{C,pd}(\Phi)$. Then
$\neg\varphi_X$ is true at some state in some awareness structure in
$\N_n^{C,pd}(\Phi)$. By part (b) of Theorem \ref{thm:equivalent},
$\neg\varphi$ is also true at some state in some HMS structure in
$\H_n^{C}(\Phi)$, a contradiction since $\varphi$ is weakly valid in
$\H_n^{C}(\Phi)$. \eprf}

\othm{thm:main} Let $\C$ be a (possibly empty) subset of
$\{\mathrm{T}', 4', 5'\}$ and let $C$ be the corresponding subset
of $\{r, t, e\}$.  Then $\AXKhran \union \C$ is a sound and
complete axiomatization of the language $\LKIn(\Phi)$ with
respect to $\H_n^C(\Phi)$. \eothm

\prf Soundness is straightforward, as usual, by induction on the
length of the proof (after showing that all the axioms are sound
and that the inference rules preserve strong validity).  We leave
details to the reader.

To prove completeness, we first define a {\em simplified HMS\/}
structure for $n$ agents to be a tuple $M = (\Sigma,\K_1, \ldots,
\K_n,\pi)$.  That is, a simplified HMS structure is an HMS structure
without the projection functions.  The definition of $\sat$
for simplified HMS structures is the same as that for HMS structures.
(Recall that the projections functions are not needed for defining
$\sat$.) Let $\H_n^-(\Phi)$ consist of all simplified HMS structures for
$n$ agents over $\Phi$ that satisfy confinedness.

\lem\label{simplified} $\AXKhran$ is a sound and complete
axiomatization of the language $\LKIn(\Phi)$ with respect to
$\H_n^-(\Phi)$. \elem

\prf Again, soundness is obvious.
For completeness, it clearly suffices to show that every $\AXKhran$-consistent
formula is satisfiable in some structure in $\H_n^-$.  As usual, we do
this by constructing a canonical model $M^c \in \H_n^-$ and showing that
every  $\AXKhran$-consistent formula of the form $\phi = i$
for $i \in \{0, 1/2, 1\}$ is satisfiable in some state of $M^c$.
Since $(M^c,s)\sat\phi$ iff $(M^c,s)\sat\phi=1$, this clearly
suffices to prove the result.

%
%


Let $M^{c}= (\Sigma^c,{\cal
K}_1^c,...,{\cal K}_n^c,\pi^c)$, where

\begin{itemize}
\item $S_{\Psi}^c=\{s_V:V$ is a maximal $\AXKhran$-consistent subset
$D_2$ and for all $p\in (\Phi-\Psi),\ p=1/2\in V$, and for all
$p\in\Psi$, $(p=1/2)\notin V\}$;

\item $\Sigma^c= \union_{\Psi\subseteq\Phi}S_{\Psi}^c$;

\item ${\cal K}_i^c(s_V)= \{s_W:V/ K_i\subseteq W\}$,
where $V/K_i = \{\varphi=1: K_i\varphi= 1\in V\}$;

\item $
\pi^c(s_V,p)= \left\{
\begin{array}{ll}
                                1 & \mbox{if $p= 1\in V$} \\
                                0 & \mbox{if $p= 0\in V$} \\
                                1/2 & \mbox{if $p= 1/2\in V$}.
                                \end{array}
                            \right.
                            $
\end{itemize}
Note that, by Lemma \ref{properlemma}(2), the interpretation $\pi^c$
is well defined.


We want to show that
%
\begin{eqnarray}
\label{eq:consiffsatis}
(M^{c},s_V)\sat\psi= j\mbox{\ \ iff\ \ }\psi= j\in V\mbox{, for
}j\in\{0,1/2,1\}.
\end{eqnarray}
%


We show that (\ref{eq:consiffsatis}) holds by
induction on the structure of formulas. If $\psi$ is a primitive
proposition, this follows from the definition of
$\pi^c(s_V,\psi)$. If $\psi= \neg\phi$ or $\psi=
\phi_1\land\phi_2$ or $\psi= \phi_1\hra\phi_2$, the argument is
similar to that of Lemma \ref{lemconsat}, we omit details here.

If $\psi= K_i\varphi$, then by the definition of $V / K_i$, if
$\psi= 1\in V$, then $\varphi=1\in V/K_i$, which implies that if
$s_W\in{\cal K}_i^c(s_V)$, then $\varphi= 1\in W$. Moreover, by
axiom B1, $\varphi=1/2\notin V$. By the induction hypothesis, this
implies that $(M^c,s_V)\sat\neg(\varphi=1/2)$ and that
$(M^{c},s_W)\sat\varphi$ for all $W$ such that $s_W\in {\cal
K}_i^c(s_V)$. This in turn implies that $(M^{c},s_V)\sat
K_i\varphi$. Thus, $(M^{c},s_V)\sat (K_i\varphi)=1$, i.e.,
$(M^{c},s_V)\sat\psi= 1$.

For the other direction, the argument is essentially identical to
analogous arguments for Kripke structures. Suppose that
$(M^{c},s_V)\sat(K_i\varphi)= 1$. It follows that the set $(V /
K_i)\cup\{\neg(\varphi= 1)\}$ is not $\AXKhran$-consistent. For
suppose otherwise. By Lemma \ref{maximlemma}, there would be a
maximal $\AXKhran$-consistent set $W$ that contains $(V /
K_i)\cup\{\neg(\varphi= 1)\}$ and, by construction, we would have
$s_W\in{\cal K}_i^c(s_V)$. By the induction hypothesis,
$(M^{c},s_W)\not\sat (\varphi= 1)$, and so
$(M^{c},s_W)\not\sat\varphi$. Thus, $(M^{c},s_V)\not\sat
K_i\varphi$, contradicting our assumption. Since $(V /
K_i)\cup\{\neg(\varphi= 1)\}$ is not $\AXKhran$-consistent, there
must be some finite subset, say $\{\phi_1,...,\phi_k,\neg(\varphi=
1)\}$, which is not $\AXKhran$-consistent. By classical
propositional reasoning (which can be applied since all formulas are
in $D_2$),
$$\AXKhran\vdash\phi_1\hra(\phi_2\hra(...\hra(\phi_k\hra( \varphi=
1)))).$$
%
By Gen,
%
\begin{equation}
\label{eqnKcons1} \AXKhran\vdash K_i
(\phi_1\hra(\phi_2\hra(...\hra(\phi_k\hra(\varphi=
1))))).\end{equation}
%
Using axiom K$'$ and classical propositional reasoning, we can show by
induction on $k$ that
\begin{eqnarray}
\label{eqnKcons2}\AXKhran &\vdash
&K_i(\phi_1\hra(\phi_2\hra(...\hra(\phi_k\hra(\varphi= 1)))))\hra
\\ \nonumber  &&\hra(K_i\phi_1\hra(K_i\phi_2\hra(...\hra(
K_i\phi_k\hra(K_i(\varphi= 1)))))).
\end{eqnarray}
%
Now by MP$'$ and Equations (\ref{eqnKcons1}) and
(\ref{eqnKcons2}), we get
$$\AXKhran\vdash(K_i\phi_1\hra(K_i\phi_2\hra(...\hra(K_i\phi_k\hra(K_i(\varphi=
1)))))).$$
\commentout{
By Prop$'$ and MP$'$, it follows that
$$\AXKhran\vdash(K_i\phi_1\hra(K_i\phi_2\hra(...\hra(K_i\phi_k\hra(K_i(\varphi=
1))))))=1.$$
%
By R1, it follows that
$$\AXKhran\vdash K_i\phi_1\hra(K_i\phi_2\hra(...\hra(K_i\phi_k\hra(K_i(\varphi=
1))))).$$
}
By Lemma \ref{properlemma}(10), it follows that
$$K_i\phi_1\hra(K_i\phi_2\hra(...\hra(K_i\phi_k\hra(K_i(\varphi=
1)))))\in V.$$
%
%
%
Since $\phi_1,...,\phi_k\in V / K_i$, there exist formulas
$\alpha_1,\ldots,\alpha_k$ such that $\phi_i$ has the form
$\alpha_i=1$, for $i=1,\ldots,k$.
By definition of $V /
K_i$, $(K_i\alpha_1)= 1,...,(K_i\alpha_k)= 1\in V$.
\commentout{
Since $(K_i \alpha_j)= 1\in V$, by Lemma
\ref{properlemma}(8,10), $(K_i (\alpha_j=1))=1\in V$, i.e., $(K_i
\phi_j)=1\in V$. Since $K_i \phi_j\in D_2$, by Prop$'$, it follows
that $K_i\phi_j\in V$ for $j\in \{1,\ldots,k\}$. We now show a
property analogous to Lemma \ref{properlemma}(3), saying that if $V$
is a maximal $\AXKhran$-consistent subset of $D_2$, and
$\phi_1,\phi_1\hra\phi_2\in V$, then $\phi_2\in V$. This follows
easily from the fact that Prop$'$ implies that $\AXKhran\vdash
\phi'\dhra(\phi'=1)$ for $\phi'\in D_2$ and Lemma
\ref{properlemma}(8,10). Then applying this property, repeatedly, we
get that $K_i (\varphi= 1)\in V$.
\alpha_j\hra(\alpha_j=1)$.  By Gen, K$'$, Prop$'$, and MP$'$,
$\AXKhran\vdash K_i\alpha_j\hra K_i(\alpha_j=1)$.}
Note that, by Prop$'$, $\AXKhran \vdash
\alpha_j\hra(\alpha_j=1)$.  So, by Gen, K$'$, Prop$'$, and MP$'$,
 $\AXKhran \vdash K_i \alpha_j \hra K_i(\alpha_j=1)$. Thus, $K_i
\alpha_j \hra K_i(\alpha_j=1) \in V$. Another application of Prop$'$
gives that $\AXKhran \vdash ((K_i \alpha_j) =1 \land (K_i \alpha_j
\hra K_i(\alpha_j=1)) \hra K_i (\alpha_j =1)$. Since $(K_i \alpha_j)
=1 \land (K_i \alpha_j \hra K_i(\alpha_j=1) \in V$, it follows that
$K_i(\alpha_j=1) \in V$; i.e., $K_i \phi_j \in V$. Note that, by
Prop$'$, $\AXKhran \vdash (\beta \land (\beta \hra \gamma)) \hra
\gamma$.  By repeatedly applying this observation and
Lemma~\ref{properlemma}(3), we get that $K_i (\phi = 1) \in V$.
Since, $(M^c,s_V)\sat (K_i\varphi)=1$ implies $(M^c,s_V)\not\sat
\varphi=1/2$, it follows by the induction hypothesis that
$\varphi=1/2\notin V$. Therefore $(\varphi=0\lor\varphi=1)\in V$, so
by axiom B2 and Lemma \ref{properlemma}(3), $(K_i \varphi)= 1\in V$,
as desired.\footnote{This proof is almost identical to the standard
modal logic proof that $(M,s_V)\sat K_i\varphi$ implies
$K_i\varphi\in V$ \cite{FHMV}.}

Finally, by axiom B1 and Lemma \ref{properlemma}(3), $(K_i\varphi)=
1/2\in V$ iff $\varphi= 1/2\in V$. By the induction hypothesis,
$\varphi= 1/2\in V$ iff
$(M^{c},s_V)\sat \varphi= 1/2$.  By the definition of $\sat$,
$(M^{c},s_V)\sat \varphi = 1/2$ iff $(M^{c},s_V)\sat K_i \phi =
1/2$.

This completes the proof of (\ref{eq:consiffsatis}). Since every
$\AXKhran$-consistent formula $\phi$ is in some maximal
$\AXKhran$-consistent set, $\phi$ must be satisfied at some state
in $M^c$.

It remains to show that $M^c$ satisfies confinedness.
So suppose that $s_V \in S_\Psi$.  We must show that $\K_i^c(s_V)
\subseteq S_{\Psi'}$ for some $\Psi' \subseteq \Psi$.  This is
equivalent to showing that, for all $s_W, s_{W'} \in \K_i^c(s_V)$
and all primitive propositions $p$, (a) $(M^c,s_W) \sat p=1/2$ iff
$(M^c,s_{W'}) \sat p=1/2$ and (b) if $(M^c,s_V) \sat p=1/2$, then
$(M^c,s_W) \sat p=1/2$.  For (a), suppose that $s_W, s_{W'} \in
\K_i(s)$ and $(M^c,s_W) \sat p=1/2$. Since $s_W \in \K_i^c(s_V)$, we
must have $(M^c,s_V) \sat \neg K_i((p \lor \neg p) = 1)$.  $V$
contains every instance of Conf2.  Thus, by (\ref{eq:consiffsatis}),
$(M^c,s_V) \sat \neg K_i(p=1/2) \hra K_i((p \lor \neg p) = 1)$.  It
follows that $(M^c,s_V) \sat K_i(p=1/2)$.  Thus, $(M^c,s_{W'}) \sat
p = 1/2$, as desired. For (b), suppose that $(M^c,s_V) \sat p =
1/2$. Since $V$ contains every instance of Conf1, it follows from
(\ref{eq:consiffsatis}) that $(M^c,s_V) \sat p=1/2 \hra K_i(p=1/2)$.
It easily follows that $(M^c,s_W) \sat p = 1/2$. Thus, $M^c \in
\H_n^-$, as desired.

To finish the proof that $\AXKhran$ is complete with respect to
$\H_n^-(\Phi)$, suppose that $\varphi$ is valid in $\H_n^-(\Phi)$.
This implies that $(\varphi= 0)\lor(\varphi=1/2)$ is not
satisfiable, so by (\ref{eq:consiffsatis}), $(\varphi=
0)\lor(\varphi=1/2)$ is not $\AXKhran$-consistent. Thus,
$\AXKhran\vdash\neg((\varphi= 0)\lor(\varphi=1/2))$. By
Prop$'$ and MP$'$, it follows that $\AXKhran\vdash\varphi= 1$ and
$\AXKhran\vdash\varphi$, as desired. \eprf

We now want to show that there exist projection functions
$\rho^c_{\Psi',\Psi}$ such that
$(\Sigma^c,\K_1^c, \ldots, \K_n^c,\pi^c,
\{\rho^c_{\Psi',\Psi}: \Psi \subseteq \Psi' \subseteq \Phi\} ) \in
\H_n(\Phi)$.
The intention is to define $\rho_{\Psi',\Psi}^c$ so that
$\rho_{\Psi',\Psi}^c(s_V) = s_W$, where $s_W \in S_\Psi^c$ and
agrees with $s_V$ on all formulas in $\LKIn(\Psi)$. (We say that
$s_V$ agrees with $s_W$ on $\phi$ if $(M^c,s_V) \sat \phi$ iff
$(M^c,s_W) \sat \phi$.) But first we must show that this is
well-defined; that is, that there exists a unique $W$ with these
properties. To this end, let $R_{\Psi',\Psi}$ be a binary relation
on states in $\Sigma^c$ such that $R_{\Psi',\Psi}(s_V,s_W)$ holds if $s_V
\in S_{\Psi'}^c$, $s_W \in S_\Psi^c$, and $s_V$ and $s_W$ agree on
formulas in $\LKIn(\Psi)$.  We want to show that $R_{\Psi',\Psi}$
actually defines a function; that is, for each state $s_V \in
S_{\Psi'}^c$, there exists a unique $s_W \in S_\Psi^c$ such that
$R_{\Psi',\Psi}(s_V,s_W)$.
The following lemma proves existence.

\lem\label{lem:exist} If $\Psi \subseteq \Psi'$, then for all $s_V
\in S_{\Psi'}^c$, there exists $s_W \in S_\Psi^c$ such that
$R_{\Psi',\Psi}(s_V,s_W)$ holds. \elem

\prf Suppose that $s_V \in S_{\Psi'}^c$. Let $V_{\Psi}$ be the
subset of $V$ containing all formulas of the form $\varphi=1$, where
$\varphi$ contains only primitive propositions in $\Psi$. It is
easily seen that $V_{\Psi}\union \{p=1/2:p\notin\Psi\}$ is
$\AXKhran$-consistent. For suppose, by way of contradiction, that it
is not $\AXKhran$-consistent. So, without loss of generality, there
exists a formula $\psi$ such that $\psi=1\in V_{\Psi}$ and
$\AXKhran\vdash
p_1=1/2\hra(p_2=1/2\hra(\ldots\hra(p_k=1/2\hra\neg(\psi=1))))$,
where $p_i\ne p_j$ for $i\ne j$ and $p_1,\ldots,p_k\in(\Phi-\Psi)$.
By Lemma \ref{simplified}, it follows that $\H_n^-\sat
p_1=1/2\hra(p_2=1/2\hra(\ldots\hra(p_k=1/2\hra\neg(\psi=1))))$.
It easily follows that $\H_n^-\sat \neg(\psi=1)$. Applying
Lemma \ref{simplified} again, we get that $\AXKhran\vdash\neg(\psi=1)$.
This is a contradiction, since $\psi=1\in V$ and $V$ is a maximal
$\AXKhran$-consistent subset of $D_2$. It follows that
$V_{\Psi}\union \{p=1/2:p\notin\Psi\}$ is contained in some maximal
$\AXKhran$-consistent subset $W$ of $D_2$. So $s_V$ and $s_W$ agrees
on all formulas of the form $\varphi=1$ for $\varphi\in\LKIn(\Psi)$
and therefore agree on all formulas in $\LKIn(\Psi)$, i.e.,
$R_{\Psi',\Psi}(s_V,s_W)$ holds. \eprf

The next lemma proves uniqueness. \lem\label{lem:unique} If $\Psi
\subseteq \Psi'$, then for all $s_V \in S_{\Psi'}^c$, $s_W, s_{W'}
\in S_{\Psi}^c$, if $R_{\Psi',\Psi}(s_V,s_W)$ and
$R_{\Psi',\Psi}(s_V,s_{W'})$ both hold, then $W = W'$. \elem

\prf Suppose that $R_{\Psi',\Psi}(s_V,s_W)$ and
$R_{\Psi',\Psi}(s_V,s_{W'})$ both hold.   We want to show that $W =
W'$.

Define a formula $\psi$ to be \emph{simple} if it is a Boolean
combination of formulas of the form $\phi = k$, where $\phi$ is
implication-free. It is easy to check that if $\phi$ is
implication-free, since $s_W \in S_\Psi$,  then $(M^c,s_W) \sat
\phi=1/2$ iff $\Phi_\phi - \Psi \ne \emptyset$; the same is true
for $s_{W'}$.  Moreover, if $\Phi_\phi \subseteq \Psi$, then $s_W$
and $s_{W'}$ agree on $\phi$.  Thus, it easily follows that $s_W$
and $s_{W'}$ agree on all simple formulas. We show that $W = W'$
by showing that every formula is equivalent to a simple formula;
that is, for every formula $\phi \in D_2$, there exists a simple
formula $\phi'$ such that $\H_n^- \sat \phi \dhra \phi'$.

First, we prove this for formulas $\phi$ of the form $\psi = k$, by
induction on the structure of $\psi$.  If
$\psi$ is a primitive proposition $p$, then
$\varphi$ is simple.
The argument is straightforward, using the
semantic definitions, if $\psi$ is of the form $\neg \psi'$,
$\psi_1 \land \psi_2$, or $\psi_1 \hra \psi_2$.

If $\psi$ has the form $K_i \psi'$, we proceed by cases.  If
$k=1/2$, then the result follows immediately from the induction
hypothesis, using the observation that $\H_n^- \sat K_i \psi' = 1/2
\dhra \psi' = 1/2$.  To deal with the case $k = 1$, for $\Phi'
\subseteq \Phi_{\psi}$, define $\sigma_{\psi,\Phi'} = \band_{p \in
\Phi'} ((p \lor \neg p) = 1) \land \band_{p \in (\Phi_{\psi}-\Phi')}
p = 1/2$.
%
By the
induction hypothesis, $\psi'=1$ is equivalent to a simple formula
$\psi''$.  Moreover, $\psi''$ is equivalent to $\bor_{\Phi'
\subseteq \Phi_{\psi}} (\psi'' \land \sigma_{\psi,\Phi'})$. Finally,
note that $\psi'' \land \sigma_{\psi,\Phi'}$ is equivalent to a
formula where each subformula $\xi = k$ of $\psi''$ such that
$\Phi_\xi - \Phi' \ne \emptyset$ is replaced by $\top$ if $k = 1/2$
and replaced by $\bot$ if $k \ne 1/2$; each subformula of the form
$\xi = 1/2$ such that $\Phi_\xi \subseteq \Phi'$ is replaced by
$\bot$. Thus, $\psi'' \land \sigma_{\psi,\Phi'}$ is equivalent to a
formula of the form $\psi_{\Phi'} \land \sigma_{\psi,\Phi'}$, where
$\psi_{\Phi'}$ is simple, all of its primitive propositions are in
$\Phi'$, and all of its subformulas have the form $\xi=0$ or
$\xi=1$.  Let $\sigma_{\psi,\Phi'}^+ = \band_{p \in \Phi'} ((p \lor
\neg p) = 1)$ and let $\sigma_{\psi,\Phi'}^- = \band_{p \in
(\Phi_\psi-\Phi')} (p = 1/2)$ (so that $\sigma_{\psi,\Phi'} =
\sigma_{\psi,\Phi'}^+ \land \sigma_{\psi,\Phi'}^-$). An easy
induction on the structure of a formula shows that $\psi_{\Phi'}
\land \sigma_{\psi,\Phi'}^+$ is equivalent to
$\xi'=1\land\sigma_{\psi,\Phi'}^+$, where $\xi'$ is an
implication-free formula. Finally, it is easy to see that
$\xi'=1\land\sigma_{\psi,\Phi'}^+$ is equivalent to a formula
$\xi''_{\Phi'}=1$, where $\xi''_{\Phi'}$ is implication-free. To
summarize, we have
$$
\H_n^- \sat \psi'=1 \dhra \bor_{\Phi' \subseteq \Phi_\psi}
(\xi''_{\Phi'}=1 \land \sigma_{\psi,\Phi'}^-).
$$
It easily follows that we have
\begin{equation}\label{eq1}
\H_n^- \sat K_i (\psi'=1) \dhra K_i \left(\bor_{\Phi' \subseteq
\Phi} (\xi''_{\Phi'}=1 \land \sigma_{\psi,\Phi'}^-)\right).
\end{equation}
It follows from confinedness that
\begin{equation}\label{eq2}
\H_n^- \sat K_i \left(\bor_{\Phi' \subseteq \Phi} (\xi''_{\Phi'}=1
\land \sigma_{\psi,\Phi'}^-)\right) \dhra \bor_{\Phi' \subseteq
\Phi} K_i (\xi''_{\Phi'}=1 \land \sigma_{\psi,\Phi'}^-).
\end{equation}
Since $\H_n^- \sat K_i (\psi_1 \land \psi_2) \dhra K_i \psi_1 \land
K_i \psi_2$ and $\H_n^- \sat \xi = 1/2 \dhra K_i(\xi = 1/2)$, it
follows that
\begin{equation}\label{eq3}
\H_n^- \sat K_i (\xi''_{\Phi'}=1 \land \sigma_{\psi,\Phi'}^-) \dhra
K_i (\xi''_{\Phi'}=1) \land \sigma_{\psi,\Phi'}^-.
\end{equation}
\commentout{Finally, note that there is an implication-free formula
 $\phi_{\Phi'}'$ such that $\H_n^- \sat \phi_{\Phi'}' = 1 \dhra
\psi_{\Phi'} \land \sigma_{\Phi'}^+)$.  We can obtain
$\phi_{\Phi'}'$ by verifying the following claim:

\clm Let $\psi_{\Phi'}$ be a Boolean combination of formulas of
the form $\xi=0$ and $\xi=1$, where $\xi$ is implication-free and
$\Phi_{\psi_{\Phi'}}\subseteq \Phi'$. Then, there is some
implication-free formula $\xi'$ such that $\psi_{\Phi'}\land
\sigma_{\Phi'}^+$ is equivalent to $\xi'=1\land
\sigma_{\Phi'}^+$.\eclm

The claim can be easily verified by induction in the structure of
$\psi_{\Phi'}$; we leave details for the reader. And $\phi_{\Phi'}'$
can be easily derived given this claim. }
%
\commentout{
 all of which are
easily seen to be valid in $\H_n^-$:
\begin{itemize}
\item $\xi_1 =1 \land \xi_2 = 1 \dhra (\xi_1 \land \xi_2) = 1$;
\item $\xi_1 =1 \lor \xi_2 = 1 \dhra (\xi_1 \lor \xi_2) = 1$;
\item $\xi=0 \dhra \neg\xi = 1$;
\item $\neg (\xi = 1) \dhra (\neg \xi = 1)$;
\item $\neg(\xi = 0) \dhra \xi = 1$;
\item $\neg (\xi_1 = k \land \xi_2 = k') \dhra \xi_i = 1-k \lor \xi_2 =
1-k'$
\item $\neg (\xi_1 = k \lor \xi_2 = k') \dhra \xi_i = 1-k \land \xi_2 =
1-k'$.
\end{itemize}
}
Finally, since $\H_n^- \sat K_i(\xi = 1) \dhra K_i \xi = 1$, we
can conclude from (\ref{eq1}), (\ref{eq2}), and (\ref{eq3}) that
$$\H_n^- \sat (K_i\psi') = 1 \dhra (K_i\xi''_{\Phi'}) = 1 \land
\sigma_{\psi,\Phi'}^-,$$ and hence $K_i \psi' = 1$ is equivalent to
a simple formula.

Since $K_i \psi = 0$ is equivalent to $\neg(K_i \psi =1) \land \neg(K_i
\psi = 1/2)$, and each of $K_i \psi = 1$ and $K_i \psi = 1/2$ is
equivalent to a simple formula, it follows that $K_i \psi = 0$ is
equivalent to a simple formula.

The arguments that $\neg \psi_1$, $\psi_1 \land \psi_2$, $\psi_1
\hra \psi_2$, and $K_i \psi_1$ are equivalent to simple formulas if
$\psi_1$ and $\psi_2$ are follows similar lines, and is left to the
reader.  It follows that every formula in $D_2$ is equivalent to a
simple formula.  This shows that $W = W'$, as desired. \eprf

It follows from Lemmas~\ref{lem:exist} and \ref{lem:unique} that
$R_{\Psi,\Psi'}$ defines a function.  We take this to be the
definition of $\rho^c_{\Psi,\Psi'}$.  We now must show that the
projection functions are coherent.

\lem If $\Psi_1 \subseteq \Psi_2 \subseteq \Psi_3$, then
$\rho_{\Psi_3,\Psi_1}^c=\rho_{\Psi_3,\Psi_2}^c\circ\rho_{\Psi_2,\Psi_1}^c$.
\elem

\prf If $s_V \in S_{\Psi_3}$, we must show that
$\rho^c_{\Psi_3,\Psi_1}(s_V)=\rho^c_{\Psi_2,\Psi_1}(\rho^c_{\Psi_3,\Psi_2}(s_V))$.
Let $s_W=\rho^c_{\Psi_3,\Psi_2}(s_V)$ and
$s_X=\rho^c_{\Psi_2,\Psi_1}(s_W)$. Then $s_W$ and $s_V$ agree on all
formulas in $\LKIn(\Psi_2)$ and $s_W$ and $s_X$ agree on all
formulas in $\LKIn(\Psi_1)$.  Thus, $s_V$ and $s_X$ agree on all
formulas in $\LKIn(\Psi_1)$.  Moreover, by construction, $s_X \in
S_{\Psi_1}$. By Lemma~\ref{lem:unique}, we must have $s_X =
\rho^c_{\Psi_3,\Psi_1}(s_V)$. \eprf

Since we have now shown that the projection functions are well defined,
from here on, we abuse notation and refer to $M^c$ as
$(\Sigma^c,\K_1^c, \ldots, \K_n^c,\pi^c,
\{\rho^c_{\Psi',\Psi}: \Psi \subseteq \Psi' \subseteq \Phi\} )$.
To complete the proof of Theorem \ref{thm:main}, we now show that
$M^c$ satisfies projection preserves knowledge and ignorance.
Both facts will follow easily from Proposition~\ref{lem:projpres} below.

We first need a lemma, which provides a condition for $s_W$ to be in
$\K_i^c(S_V)$ that is easier to check.

\lem\label{lem:inKi} If $\K_i^c(s_V) \subseteq S^c_{\Psi}$, $s_W \in
S^c_{\Psi}$, and $V/K_i \inter \LKInt(\Psi) \subseteq W$, then $s_W
\in \K_i^c(s_V)$. 

\elem \prf Suppose that $V$ and $W$ are as in the
antecedent of the statement of the lemma.  We must show that $V/K_i
\subseteq W$.

First note that $V/K_i$ is closed under implication.  That is, if
$\phi_1 = 1 \in V/K_i$, and $\AXKhran \vdash \phi_1 \hra \phi_2$,
then $\phi_2 = 1 \in V/K_i$. This follows from the observations that
if $\AXKhran \vdash \phi_1 \hra \phi_2$, then $\AXKhran \stur K_i
(\phi_1 \hra \phi_2)$, and $\AXKhran \stur K_i \phi_1 \hra K_i
\phi_2$;
so, by Prop$'$, $\AXKhran \stur (K_i \phi_1 \hra K_i \phi_2)=1$.
Thus, $(K_i \phi_1 \hra K_i \phi_2)=1 \in V$.  Moreover, since $\phi_1 =
1\in V/K_i$, we must have $K_i \phi_1 = 1 \in V$.  Finally, by Prop$'$,
$\AXKhran \stur (K_i \phi_1 = 1 \land (K_i \phi_1 \hra K_i \phi_2)=1)
\hra K_i \phi_2 = 1$.  Thus, Lemma \ref{properlemma}(3), 
$K_i \phi_2 = 1\in V$.  So $\phi_2 = 1 \in W/K_i$, as desired.
By Lemma \ref{properlemma}(3), $W$ is also closed under implication.  
Thus, by the proof of Lemma~\ref{lem:unique}, it suffices to show
that
$\phi=1 \in W$ for each simple formula $\phi$ such that $K_i \phi =
1 \in V$. Indeed, we can take $\phi$ to be in conjunctive normal
form: a conjunction of disjunctions of formulas of \emph{basic
formulas}, that is formulas of the form $\psi=k$ 
where $\psi$ is implication-free.  
Moreover, since 
by Lemma~\ref{properlemma}(4), 
$W$ is  closed under conjunction ($\phi_1 \in W$ and $\phi_2 \in
W$ implies that $\phi_1 \land \phi_2 \in W$) and it is easy to show that
$V/K_i$ is closed under breaking up of conjunctions (if $\phi_1 \land
\phi_2 \in V/K_i$ then $\phi_i  \in V/K_i$ for $i=1,2$),
it suffices to show that $\psi=1
\in W$ for each disjunction $\psi$ of basic formulas such that $K_i
\psi = 1 \in V$.  We proceed by induction on the number of disjuncts
in $\psi$.

If there is only one disjunct in $\psi$, that is, $\psi$ has the
form $\psi' = k$, where $\psi'$ is implication-free, suppose first
that $k=1/2$.  It is easy to check that $\H_n^- \sat (\psi' = 1/2)
\dhra \bor_{p \in \Phi_{\psi'}} (p = 1/2)$ and $\H_n^- \sat
K_i(\psi' = 1/2) \dhra \bor_{p \in \Phi_{\psi'}} K_i(p = 1/2)$.
Since $(K_i(\psi' = 1/2))=1 \in V$, by Prop$'$ and Lemma
\ref{properlemma}(3) $K_i(\psi' = 1/2) \in V$ and $K_i(p=1/2) \in V$
for some $p \in \Phi_{\psi'}$ (since, by Lemma~\ref{simplified}, for
all formulas $\sigma$, we have $\sigma \in V$ iff $(M^c,s_V) \sat
\sigma$).
Since $\K_i^c(s_V) \subseteq S^c_\Psi$, it must be the case that
$K_i(p = 1/2) \in V$ iff $p \notin \Psi$.  Since $s_W \in
S^c_{\Psi}$, $p = 1/2 \in W$.  Since $W$ is closed under
implication, $\psi' = 1/2 \in W$, as
desired. If $k = 0$ or $k = 1$, then it is easy to see that $\psi =
1 \in V/K_i$ only if $\psi=1 \in \LKInt(\Psi)$ so, by assumption,
$\psi = 1 \in W$.

If $\psi$ has more than one disjunct, suppose that there is some
disjunct of the form $\psi' = 1/2$ in $\psi$.   If there is some $p
\in (\Phi_{\psi'} -\Psi)$ then, as above, we have $K_i(p = 1/2) \in
V$ and $p = 1/2 \in W$, and, thus, $\psi'=1/2 \in W$. Therefore,
$\psi=1\in W$.
If there is no primitive proposition $p \in (\Phi_{\psi'} -\Psi)$,
then $(M^c,s_V) \sat K_i(\psi' \ne 1/2) = 1$, and thus $(\psi' \ne
1/2) = 1 \in V/K_i$.   It follows that if $\psi''$ is the formula
that results from removing the disjunct $\psi' = 1/2$ from $\psi$,
then $\psi'' \in V/K_i$.  The result now follows from the induction
hypothesis.
Thus, we can assume without loss of generality that every disjunct of
$\psi$ has the form $\psi' = 0$ or $\psi' =1$.  If there is some
disjunct $\psi' = k$, $k \in \{0,1\}$, that mentions a primitive
proposition $p$ such that $p \notin \Psi$, then it is easy to check
that $K_i(\psi' = 1/2) \in V$.  Thus, $(\psi' = 1/2) = 1 \in V/K_i$.
Again, it follows that if $\psi''$ is the formula that results from
removing the disjunct $\psi' = k$ from $\psi$, then $\psi'' \in
V/K_i$ and, again, the result follows from the induction hypothesis.
Thus, we can assume that $\psi \in \LKInt(\Psi)$. But then $\psi \in
V/K_i$, by assumption.  \eprf

\pro\label{lem:projpres} Suppose $\Psi_1\subseteq\Psi_2$,
$s_V\in S^c_{\Psi_2}$, $s_W=\rho_{\Psi_2,\Psi_1}^{c}(s_V)$,
$\K_i^c(s_V)\subseteq S_{\Psi_3}^c$, and $\K_i^c(s_W)\subseteq
S_{\Psi_4}^c$. Then $\Psi_4 = \Psi_1 \inter \Psi_3$ and
$\rho^c_{\Psi_3,\Psi_4}(\K_i^c(s_V))=\K_i^c(s_W)$.
\epro

\prf By the definition of projection, $\Psi_4 \subseteq \Psi_1$.
To show that $\Psi_4\subseteq\Psi_3$, suppose that
$p\in\Psi_4$.  Since $\K_i^c(s_W) \subseteq S^c_{\Psi_4}$,
$(M^c,s_W)\sat K_i(p\lor\neg p)$.  Since  $\Psi_4\subseteq\Psi_1$,
by Lemma~\ref{lem:moredefined}, $(M^c,s_V)\sat K_i(p\lor\neg p)$. Thus,
$(M^c,s_{V'})\sat p\lor\neg p$ for all $s_{V'}\in
\K_i^c(s_V)$. Therefore, $p\in\Psi_3$.
Thus, $\Psi_4 \subseteq \Psi_1 \inter \Psi_3$.  For the opposite
containment, if $p \in \Psi_1 \inter \Psi_3$, then $(M^c,s_V)\sat
K_i(p\lor\neg p)$. By definition of projection, since $p \in \Psi_1$ and
$s_W=\rho_{\Psi_2,\Psi_1}^{c}(s_V)$, $(M^c,s_W)\sat K_i(p\lor\neg p)$.
Thus, $p \in \Psi_4$.  It follows that $\Psi_4 = \Psi_1 \inter \Psi_3$.

To show that $\rho^c_{\Psi_3,\Psi_4}(\K_i^c(s_V))=\K_i^c(s_W)$, we
first prove that
$\rho^c_{\Psi_3,\Psi_4}(\K_i^c(s_V))\supseteq\K_i^c(s_W)$.
Suppose that $s_{W'}\in\K_i^c(s_W)$. We construct $V'$ such that
$\rho^{c}_{\Psi_3,\Psi_4}(s_{V'})=s_{W'}$ and
$s_{V'}\in\K_i^c(s_V)$. We claim that $V/K_i\cup\{\varphi=
1:\varphi= 1\in W', \varphi\mbox{ is implication-free}\}$ is
$\AXKhran$-consistent. For suppose not. Then there exist
formulas $\varphi_1,...,\varphi_m,\varphi'_1,...,\varphi'_k$ such that
$\varphi_j= 1\in V/K_i$ for $j\in\{1,...,m\}$, $\varphi'_j$ is
implication-free and $\varphi'_j= 1\in W'$ for $j\in\{1,...,k\}$,
and $\{\varphi_1= 1,...,\varphi_m= 1,\varphi'_1= 1,...,\varphi'_k=
1\}$ is not $\AXKhran$-consistent. Let $\psi=
\varphi'_1\land...\land\varphi'_k$. Then $\psi= 1\in W'$, so
$\psi=1/2\not\in W$. Thus, by axiom B1 and Lemma
\ref{properlemma}(3), $(K_i\neg\psi)=1/2\not\in W$. By definition of
${\cal K}_i^c$, $(K_i\neg\psi)=1\not\in W$, for otherwise
$\neg\psi=1\in W'$. By axiom P11, it follows that
$(K_i\neg\psi)=0\in W$. Thus, by Lemma \ref{properlemma}(7), $(\neg
K_i\neg\psi)= 1\in W$, so $(\neg K_i\neg\psi)= 1\in V$. Thus, there
must be some maximal $\AXKhran$-consistent set $V''\supseteq V/K_i$
such that $(\neg\psi)= 1\notin V''$. Since
$s_{V''}\in\K_i^c(s_V)\subseteq S_{\Psi_3}^c$,
$\Psi_4\subseteq\Psi_3$, $s_{W'}\in S_{\Psi_4}$, and $(\neg\psi)=
1/2\notin W'$, we have
$(\neg\psi)= 1/2\notin V''$. So, by axiom P11, $(\neg\psi)= 0\in
V''$. Thus, $\psi=1\in V''$ and $\varphi'_1= 1,...,\varphi'_k= 1\in
V''$, which is a
contradiction since $V''\supseteq V/K_i$ and $V''$ is
$\AXKhran$-consistent. Let $V'$ be an $\AXKhran$-consistent set
containing $V/K_i\cup\{\varphi= 1:\varphi= 1\in W', \varphi\mbox{ is
implication-free}\}$. By construction, $s_{V'}\in\K_i^c(s_V)$.
Moreover, since $(\neg\varphi)=1\dhra\varphi=0$ is valid in $\H_n$
and $s_{W'}\in S_{\Psi_4}^c$, $\rho^c_{\Psi_3,\Psi_4}(s_{V'})$
agrees with $s_{W'}$ on all formulas of the form $\varphi=k$ for
$k\in\{0,1/2,1\}$ and $\varphi$ implication-free. Therefore, it is
easy to show using Prop$'$ that they agree on all simple formulas.
Thus, by the proof of Lemma \ref{lem:unique}, they agree on all
formulas in $\LKIn(\Psi_4)$. By uniqueness of $\rho^c$, it follows
that $\rho^c_{\Psi_3,\Psi_4}(s_{V'})=s_{W'}$, as desired.

\commentout{
To prove that
$\rho^c_{\Psi_3,\Psi_4}(\K_i^c(s_V))\subseteq\K_i^c(s_W)$, suppose
that $s_{V'} \in \K_i^c(s_V)$ and $s_{W'} =
\rho^c_{\Psi_3,\Psi_4}(s_{V'})$.  By Lemma~\ref{lem:inKi}, it
suffices to show  that $s_W/K_i \inter \LKInt(\Psi_4) \subseteq
S_{W'}$. But if $\phi = 1 \in S_W/K_i \inter \LKInt(\Psi_4)$, then
$K_i \phi = 1 \in S_W$, and so $K_i \phi = 1 \in S_V$.  Thus, $\phi
=1 \in S_{V'}$. Since $s_{W'} = \rho^c_{\Psi_3,\Psi_4}(s_{V'})$, we
must have $\phi = 1 \in S_{W'}$, as desired.
}

The proof of the other direction
$\rho^c_{\Psi_3,\Psi_4}(\K_i^c(s_V))\subseteq\K_i^c(s_W)$ is
similar. Suppose that $s_{V'}\in\K_i^c(s_V)$ and
$s_{V''}=\rho^c_{\Psi_3,\Psi_4}(s_{V'})$. We need to prove that
$s_{V''}\in\K_i^c(S_W)$. We claim that $W/K_i\cup\{\varphi=
1:\varphi= 1\in V'', \varphi\mbox{ is implication-free}\}$ is
$\AXKhran$-consistent. For suppose not. Then there exist formulas
$\varphi_1,...,\varphi_m,\varphi'_1,...,\varphi'_k$ such that
$\varphi_j= 1\in W/K_i$ for $j\in\{1,...,m\}$, $\varphi'_j$ is
implication-free and $\varphi'_j= 1\in V''$ for $j\in\{1,...,k\}$,
and $\{\varphi_1= 1,...,\varphi_m= 1,\varphi'_1= 1,...,\varphi'_k=
1\}$ is not $\AXKhran$-consistent. Let $\psi=
\varphi'_1\land...\land\varphi'_k$. Then $\psi= 1\in V''$, so
$\psi=1/2\not\in V$. Thus, by axiom B1 and Lemma
\ref{properlemma}(3), $(K_i\neg\psi)=1/2\not\in V$.
Using Lemmas \ref{lem:moredefined} and \ref{simplified}, it is easy
to show that $\psi= 1\in V''$ implies $\psi= 1\in V'$, so, by
definition of ${\cal K}_i^c$, $(K_i\neg\psi)=1\not\in V$, for
otherwise $\neg\psi=1\in V'$. By axiom P11, it follows that
$(K_i\neg\psi)=0\in V$. Thus, by Lemma \ref{properlemma}(7), $(\neg
K_i\neg\psi)= 1\in V$. But as $s_{V''}\in S_{\Psi_4}^c$, $\psi= 1\in
V''$ implies $\Phi_{\psi}\in S_{\Psi_4}^c$. Since
$\Psi_4\subseteq\Psi_1$, it follows that $\psi= 1/2\notin W$, thus,
by definition of projection, we have $(\neg K_i\neg\psi)= 1\in W$.
Thus, there must be some maximal $\AXKhran$-consistent set
$W'\supseteq W/K_i$ such that $(\neg\psi)= 1\notin W'$. Since
$s_{W'}\in\K_i^c(s_W)\subseteq S_{\Psi_4}^c$, we have $(\neg\psi)=
1/2\notin W'$. So by P11 it follows that $(\neg\psi)= 0\in W'$, so
$\psi= 1\in W'$. Thus, $\varphi'_1= 1,...,\varphi'_k= 1\in W'$,
which is a contradiction since $W'\supseteq W/K_i$ and $W'$ is
$\AXKhran$-consistent. Let $W'$ be an $\AXKhran$-consistent set
containing $W/K_i\cup\{\varphi= 1:\varphi= 1\in V'', \varphi\mbox{
is implication-free}\}$. By construction, $s_{W'}\in\K_i^c(s_W)$.
Moreover, since $(\neg\varphi)=1\dhra\varphi=0$ is valid in $\H_n$
and $s_{W'},s_{V''}\in S_{\Psi_4}^c$, $s_{V''}$ agrees with $s_{W'}$
on all formulas of the form $\varphi=k$ for $k\in\{0,1/2,1\}$ and
$\varphi$ implication-free. Therefore, it is easy to show using
Prop$'$ that they agree on all simple formulas. Thus, by the proof
of Lemma \ref{lem:unique}, they agree on all formulas in
$\LKIn(\Psi_4)$. By uniqueness of $\rho^c$, it follows that
$s_{V''}=s_{W'}\in\K_i^c(s_W)$, as desired. \eprf

The following result is immediate from Proposition~\ref{lem:projpres}.
\cor Projection preserves knowledge and ignorance in $M^c$. \ecor

Since projections preserve knowledge and ignorance in $M^c$, it
follows that $M^c\in\H_n(\Phi)$.
This finishes the proof for the case ${\cal C}=\emptyset$. If
$T'\in{\cal C}$, we must
show that $M^c$ satisfies generalized reflexivity. Given $s_V\in
S_{\Psi_1}^{c}$, by confinedness, $\K_i^c(s_V)\subseteq
S_{\Psi_2}^{c}$ for some $\Psi_2\subseteq\Psi_1$. It clearly
suffices to show that $s_W = \rho^c_{\Psi_1,\Psi_2}(s_V)\in\K_i^c(s_V)$.
By Lemma~\ref{lem:inKi}, to prove this, it suffices to show that
$V/K_i \inter \LKInt(\Psi_2) \subseteq W$.  If $\phi=1 \in V/K_i
\inter \LKInt(\Psi_2)$, then $K_i\phi = 1 \in V$.
Note that Prop$'$ implies that $\AXKhran\vdash
\varphi\hra\varphi=1$, which by Gen, K$'$, Prop$'$, and MP$'$
implies that $\AXKhran\vdash (K_i\varphi\hra K_i(\varphi=1))=1$.
Therefore, as $(K_i \varphi)= 1\in V$, by Lemma
\ref{properlemma}(8,10), $(K_i (\varphi=1))=1\in V$. By Prop$'$ and
Lemma \ref{properlemma}(3), it follows that $K_i(\phi = 1) \in V$.
By T$'$ and Lemma \ref{properlemma}(3), it follows that $\phi=1 \lor
\bigvee_{\{p:p\in\Phi_{\varphi}\}}K_i(p= 1/2) \in V$. But since
$\phi \in \LKInt(\Psi_2)$, we must have $\Phi_{\varphi} \subseteq
\Psi_2$. Moreover, since $\K_i^c(s_V) \subseteq S^c_{\Psi_2}$, we
must have $\neg K_i (p=1/2) \in V$ for all $p \in \Psi_2$.  Thus, it
easily follows that $\phi=1 \in V$. Finally, since $s_W =
\rho^c_{\Psi_1,\Psi_2}(s_V)$, we must have $\phi=1 \in W$, as
desired.

Now suppose $4'\in{\cal C}$. We want to show that $M^c$ satisfies
part (a) of stationarity. Suppose that $s_W\in {\cal K}_i^c(s_V)$
and $s_X\in {\cal K}_i^c(S_W)$.  We must show that $s_X\in {\cal
K}_i^c(S_V)$. If $(K_i\varphi)= 1\in V$ then, by axioms Prop$'$
and 4$'$ and Lemma \ref{properlemma}(3), $(K_iK_i\varphi)= 1\in V$.
This implies that $(K_i \varphi)= 1\in W$, which implies that
$\varphi= 1\in X$. Thus, $V / K_i\subseteq X$ and $s_X\in{\cal
K}_i^c(s_V)$, as desired.

Finally, suppose that $5'\in{\cal C}$. We want to show that $M^c$
satisfies part (b) of stationarity. Suppose that $s_W\in {\cal
K}_i^c(s_V)$ and $s_X\in {\cal K}_i^c(s_V)$.  We must show that
$s_X\in {\cal K}_i^c(s_W)$.
\commentout{ Note that Prop$'$ implies that $\AXKhran\vdash
\varphi\hra\varphi=1$, which by Gen, K$'$, Prop$'$, and MP$'$
implies that $\AXKhran\vdash (K_i\varphi\hra K_i(\varphi=1))=1$.
Therefore, if $(K_i \varphi)= 1\in W$, then by Lemma
\ref{properlemma}(8,10), $(K_i (\varphi=1))=1\in W$. Since
$s_W\in{\cal K}_i^c(s_V)$, by Lemma \ref{properlemma}(6), $(K_i\neg
K_i(\varphi=1))= 1\notin V$. Since $\varphi=1\in D_2$, by Lemma
\ref{properlemma}(2), $(K_i\neg K_i(\varphi=1))= 0\in V$. By Lemma
\ref{properlemma}(7), $(\neg K_i\neg K_i(\varphi=1))= 1\in V$. So,
by axioms Prop$'$ and 5$'$ and Lemma \ref{properlemma}(8,10),
$(K_i(\varphi=1))= 1\in V$. By axiom B2 and Lemma
\ref{properlemma}(3), $(K_i\varphi)= 1\in V$. Therefore, $\varphi=
1\in X$, so $W / K_i\subseteq X$ and we have $s_X\in {\cal
K}_i^c(s_W)$, as desired.
}
Suppose by way of contradiction that $s_X\notin {\cal K}_i^c(s_W)$,
then there exists $\varphi=1\in W/K_i$ such that $\varphi=1\notin
X$. $\varphi=1\in W/K_i$ implies that $(K_i\varphi)=1\in W$, so by
Prop$'$, B1, and Lemma \ref{properlemma}(3) it follows that
$\varphi=1/2\notin W$. By Conf2 it can be easily shown that
$\varphi=1/2\notin X$, so by Prop$'$, we get that $\varphi=0\in X$.
As in the proof of the case $T'\in{\cal C}$, if $(K_i \varphi)= 1\in
W$, then $(K_i (\varphi=1))=1\in W$. Then it follows that $(K_i\neg
K_i(\varphi=1))=1\notin V$, for otherwise since $s_W\in\K_i^c(s_V)$
we get that $(\neg K_i(\varphi=1))=1\in W$. Since $(K_i\neg
K_i(\varphi=1))\in D_2$, it is easy to show that $(\neg K_i\neg
K_i(\varphi=1))=1\in V$. By Prop$'$ and 5$'$ and Lemma
\ref{properlemma}(8,10), it follows that $(K_i(\varphi=1)\lor
K_i((\varphi=1)=1/2))=1\in V$. Then using Lemma
\ref{properlemma}(5,6,7), it easily follows that either
$(K_i(\varphi=1))=1\in V$ or $(K_i((\varphi=1)=1/2))=1\in V$. Then,
either $(\varphi=1)=1\in X$ or $((\varphi=1)=1/2)=1\in X$, but this
is a contradiction since $\varphi=0\in X$ and $X$ is
$\AXKhran$-consistent. \eprf

\end{document}